\documentclass[11pt]{article}

\usepackage[final]{acl}

\usepackage{times}
\usepackage{latexsym}

\usepackage[T1]{fontenc}

\usepackage[utf8]{inputenc}

\usepackage{microtype}
\usepackage{amsmath}

\usepackage{inconsolata}
\usepackage{multirow} 
\usepackage{enumitem}
\usepackage{graphicx}
\usepackage{xspace}
\usepackage{cleveref} 
\usepackage{longtable}    
\usepackage{makecell}

\PassOptionsToPackage{nameinlink}{cleveref}
\crefformat{section}{\S#2#1#3}
\crefformat{subsection}{\S#2#1#3}
\crefformat{subsubsection}{\S#2#1#3}

\newcommand{\method}{\textsc{PACMem}\xspace}
\newcommand{\benchmark}{\textsc{PersonaTrail}\xspace}
\newcommand{\pavg}{\ensuremath{\mathcal{P}_{\text{avg}}}}

\usepackage{verbatim}
\usepackage[utf8]{inputenc}
\usepackage{amssymb}
\usepackage[most]{tcolorbox}
\usepackage{booktabs}
\usepackage[dvipsnames]{xcolor}   
\usepackage{pifont} 
\usepackage{subcaption}
\usepackage{capt-of}
\usepackage{tikz}
\usepackage{fvextra}
\newcommand{\cmark}{\textcolor{Green}{\ding{51}}}
\newcommand{\xmark}{\textcolor{red}{\ding{55}}}

\usepackage{titletoc}

\newcommand{\prompttbox}[4]{
  \begin{figure*}[htbp]
  \centering
  \begin{tcolorbox}[
    colback=blue!5!white, colframe=blue!75!black, width=0.95\textwidth,
    arc=1mm, boxrule=0.8pt, colbacktitle=blue!75!black, coltitle=white,
    fonttitle=\bfseries\sffamily\small, halign title=center,
    top=2mm, bottom=2mm, left=2mm, right=2mm,
    fontupper=\ttfamily\tiny,
    title={\textbf{\mbox{#1}}}
    
  ]
  \VerbatimInput[breaklines=true, fontsize=\tiny,breaksymbol={}]{#2}
  \end{tcolorbox}
  \caption{#3}
  \label{fig:#4}
  \end{figure*}
}
\title{\benchmark: Benchmarking Personalized Web Agents \\ through Browsing Trails}

\author{
  Seungbin Yang\textsuperscript{1}\thanks{\ Equal contribution.} \quad
  Chaewoon Ki\textsuperscript{2}\footnotemark[1] \quad
  Dohyun Lee\textsuperscript{1} \quad
  Jaegul Choo\textsuperscript{1} \quad
  ChaeHun Park\textsuperscript{3} \\
  \textsuperscript{1}KAIST AI \quad
  \textsuperscript{2}KAIST School of Computing \quad
  \textsuperscript{3}Chonnam National University \\
  \texttt{\{sby99, befilledwith, aiclaudev, jchoo\}@kaist.ac.kr} \\
  \texttt{chaehun.park@jnu.ac.kr}
}

\begin{document}
\renewcommand{\thefootnote}{\fnsymbol{footnote}}
\maketitle
\renewcommand{\thefootnote}{\arabic{footnote}}

\begin{abstract}
\label{sec:abstract}

Recent advances in large language models have enabled web agents to autonomously execute complex tasks.
In practice, users frequently provide underspecified instructions, requiring agents to infer the missing context from their raw browsing histories.
Existing benchmarks fail to capture this form of \textit{personalization}, as they either restrict tasks to fully explicit prompts or abstract web interaction history into simplified forms.
To bridge this gap, we introduce \benchmark, a benchmark for personalized web agents operating in a managed open web environment. 
By leveraging realistic browsing trajectories as user history, \benchmark evaluates an agent's ability to infer user preferences and recall information from past browsing sessions.
We further provide Preference-Aware Contextual Memory (\method), a framework that decomposes raw browsing histories into two types of structured memory: factual memories that summarize individual sessions and preference memories that distill recurring behavioral patterns.
At inference time, the agent retrieves the most relevant entries from these memories to guide personalized navigation.
Extensive experiments show that \method consistently outperforms existing memory-based baselines on both tasks.
\end{abstract}
\section{Introduction} \label{sec:introduction}

The continuous evolution of large language models (LLMs) has expanded the capacity of intelligent agents to handle intricate web navigation scenarios~\citep{yao2022webshop,deng2023mind2web,gur2024real,zhou2024webarena,zheng2024gpt}.
By interacting with live websites on behalf of users, web agents have emerged as a promising tool for automating a broad range of real-world web tasks such as online shopping and information retrieval~\citep{he2024webvoyager,pan2024webcanvas}.

\begin{figure}[t]
    \centering
        \includegraphics[width=\columnwidth]{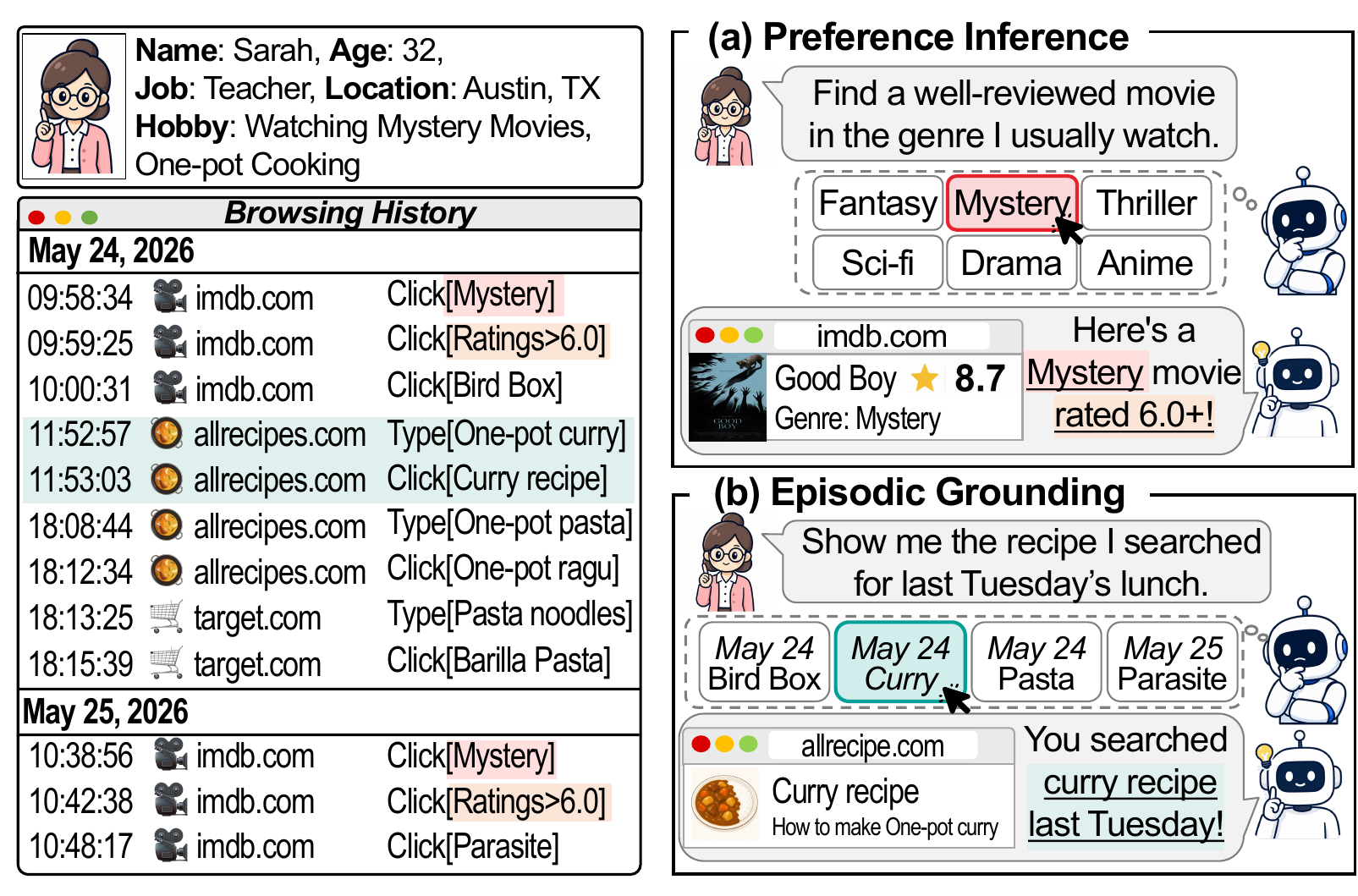}
        \vspace{-0.5cm}
        \captionsetup{skip=6pt}
    \caption{Two forms of personalized web agent queries: (a) \textit{preference inference}, where user preferences must be inferred from recurring browsing patterns, and (b) \textit{episodic grounding}, where the web agent must retrieve a specific past session referenced in the query.}
    \label{fig:intro}
    \vspace{-1.0em}
\end{figure}


Despite this practical utility, current benchmarks struggle to evaluate web agents in realistic, user-centric scenarios.
In practice, users repeatedly interact with the web, naturally accumulating raw browsing histories.
Consequently, rather than providing fully-specified instructions, users expect agents to leverage this history to resolve their under-specified queries that omit personal preferences or reference past interactions~\citep{10.1145/176789.176792, horvitz1999principles,10.1145/1076034.1076111,wan2025infer}.
However, most existing benchmarks restrict evaluations to explicit instructions that provide all necessary information upfront, failing to reflect the historical context required for personalized user queries~\citep{he2024webvoyager,pan2024webcanvas}.
While recent works have made progress toward personalization, their evaluations are confined to transaction records~\citep{cai2025large} or LLM-synthesized artificial histories~\citep{kim2026persona2web}, rather than fine-grained interaction trajectories executed on live websites.
    
\begin{table*}[t]
\centering
\setlength{\tabcolsep}{4pt}
\renewcommand{\arraystretch}{1.15}
\resizebox{\textwidth}{!}{%
\begin{tabular}{lcccccccc}
\toprule
\textbf{Benchmark} & \textbf{\# Domain} & \textbf{\# Website} & \textbf{\# Query} & \textbf{Pref.} & \textbf{Epi.} & \textbf{Multi-hop} & \textbf{User History} & \textbf{Environment} \\
\midrule
MiniWoB++~\citep{liu2018reinforcement}  & --   & 100 & 100      & \xmark & \xmark & \xmark & \xmark         & Simulated         \\
WebShop~\citep{yao2022webshop}          & 1    & 1   & 12{,}087 & \xmark & \xmark & \xmark & \xmark         & Simulated         \\
WebArena~\citep{zhou2024webarena}           & 6    & 6   & 812      & \xmark & \xmark & \cmark & \xmark         & Simulated         \\
Mind2Web~\citep{deng2023mind2web}       & 31   & 137 & 2{,}350  & \xmark & \xmark & \xmark & \xmark         & Cached            \\
WebVoyager~\citep{he2024webvoyager}     & 5    & 15  & 643      & \xmark & \xmark & \xmark & \xmark         & Open Web          \\
Mind2Web-Live~\citep{pan2024webcanvas}  & 19   & 69  & 542      & \xmark & \xmark & \xmark & \xmark         & Managed Open Web  \\
REAL~\citep{gargreal}                   & --   & 11  & 112      & \xmark & \xmark & \xmark & \xmark         & Simulated         \\
PersonalWAB~\citep{cai2025large}        & 1    & 1   & 9{,}070  & \cmark & \xmark & \xmark & Trans. Record  & Function Call     \\
Persona2Web~\citep{kim2026persona2web}  & 21   & 105 & 450       & \cmark & \xmark & \xmark & LLM-Synth.     & Open Web          \\
\midrule
\textbf{\benchmark{} (Ours)}            & 23(114$^{\dagger}$)   & 317  & 2{,}524       & \cmark & \cmark & \cmark & Trajectory     & Managed Open Web  \\
\bottomrule
\end{tabular}%
}
\captionsetup{skip=10pt}
\caption{
Comparison of \benchmark{} with existing web agent benchmarks.
$^{\dagger}$ denotes the number of subdomains.
\textbf{Pref.} and \textbf{Epi.} indicate whether the benchmark evaluates agents' ability to infer user preferences from history and recall specific past browsing episodes, respectively.
\textbf{User History} specifies how past user activity is represented: \textit{Trans. Record} for transaction records comprising product metadata and review pairs, \textit{LLM-Synth.} for LLM-synthesized sequences of abstracted action types, and \textit{Trajectory} for fine-grained browser-level navigation logs.
}
\label{tab:benchmark_comparison}
\vspace{-0.8em}
\end{table*}
To address this limitation, we introduce \benchmark, a benchmark for personalized web agents designed with four key features. \textbf{(1) Personalization tasks.} As illustrated in Figure~\ref{fig:intro}, we define two personalization tasks: \textit{preference inference}, which requires extracting latent patterns from recurring behavior to resolve under-specified queries, and \textit{episodic grounding}, which involves referencing a specific past interaction to fulfill the current request.
\textbf{(2) Realistic user history.} Our benchmark provides browser-level trajectories executed on real-world websites, offering detailed interactions across websites as user history.
\textbf{(3) Single-hop and multi-hop queries.} We design both single-site tasks and cross-site multi-hop queries, enabling a fine-grained evaluation of agent capabilities across varying complexities.
\textbf{(4) Managed open web environment.} We adopt a managed open web paradigm~\citep{pan2024webcanvas} to ensure reproducibility despite website changes.

Through evaluations on \benchmark, we find that existing memory-based web agents struggle with personalized web navigation because they discard the episode-level details and latent preferences essential for adapting to each user.
We therefore build \textbf{P}reference-\textbf{A}ware \textbf{C}ontextual \textbf{Mem}ory (\method), which organizes raw browsing histories into a two-level hierarchy: \textit{factual memory} that captures individual session episodes, and \textit{preference memory} that distills recurring behavioral patterns into generalized user schemas.
Experiments show that \method consistently outperforms existing memory-augmented baselines across diverse backbones and task complexities.
These gains stem from transforming raw histories into structured memory representations that retain episodic details and preference signals, providing a starting point for memory design in personalized web agents.

In summary, our contributions are as follows:
\begin{itemize}[itemsep=0.2mm]
\item We introduce \benchmark{}, a personalized web agent benchmark for preference inference and episodic grounding, with fine-grained live-web browsing trajectories as user history.
\item We provide \method, a simple and effective reference method that transforms browser-level histories into structured factual and preference memories, establishing a strong baseline for personalized web navigation.
\item Extensive experiments show that 
distilling raw browsing trajectories into episodic events and latent preferences enables \method to outperform memory-augmented baselines.
\end{itemize}
\section{Related work} \label{sec:related_work}
\noindent{\textbf{Web Agent Benchmarks.}}
The evaluation of web agents has progressed from simulated or sandboxed environments~\citep{liu2018reinforcement,yao2022webshop,zhou2024webarena,koh2024visualwebarena} to cached datasets over real websites~\citep{deng2023mind2web}, culminating in live web assessments~\citep{he2024webvoyager,pan2024webcanvas,zheng2024gpt}.
However, these benchmarks do not account for personalized queries that require referencing user history.
To fill this gap, \citet{cai2025large} introduce PersonalWAB, which studies personalization in a single-domain, constrained function-call-based environment. 
Persona2Web~\citep{kim2026persona2web} extends this line of work to open web tasks with LLM-synthesized histories based on predefined action types. 
However, these works represent user history through transaction records or synthesized high-level events, rather than browser-level interaction trajectories.
In reality, user preferences and episodic contexts are embedded within low-level interactions, such as clicks, searches, and page transitions.
Therefore, accurately evaluating personalized web agents requires browsing trajectories that preserve user action sequences.
In response, \benchmark{} grounds personalized evaluation by collecting interaction-level trajectories from live websites.
A detailed comparison is summarized in Table~\ref{tab:benchmark_comparison}.

\noindent{\textbf{Personalization in LLM Agents.}}
Personalizing LLM outputs by conditioning on user-specific data has been widely studied through retrieval-augmented approaches that select relevant items from user profiles or interaction histories~\citep{salemi2024lamp,salemi2024optimization,wu2024understanding,kumar2024longlamp}.
In the LLM agent domain, several works propose structured memory systems tailored for personalization: 
\citet{kang-etal-2025-memory} organize memory into hierarchical OS-inspired tiers, and \citet{xu2026mem} structure memories into self-linking note formats.
\citet{lyu2026personalalign} and \citet{wang2026me} employ hierarchical memories for GUI and mobile agents, whereas \citet{hu2026orion} utilize global-micro profiling for API-driven web navigation.
Concurrently, memory-augmented web agents like AWM~\citep{wang2025agent} and ReasoningBank~\citep{ouyang2026reasoningbank} explore how to retain and reuse past execution traces.
However, these lines of work remain only partially connected: personalized agents primarily target dialogue or mobile contexts, whereas memory-augmented web agents focus on functional task automation rather than extracting latent user preferences from long, noisy, cross-session trajectories.
\method addresses this gap by structuring web trajectories into factual and preference memories for personalized navigation.
\section{\benchmark} \label{sec:benchmark}

\begin{figure}[t]
    \centering
        \includegraphics[width=\columnwidth]{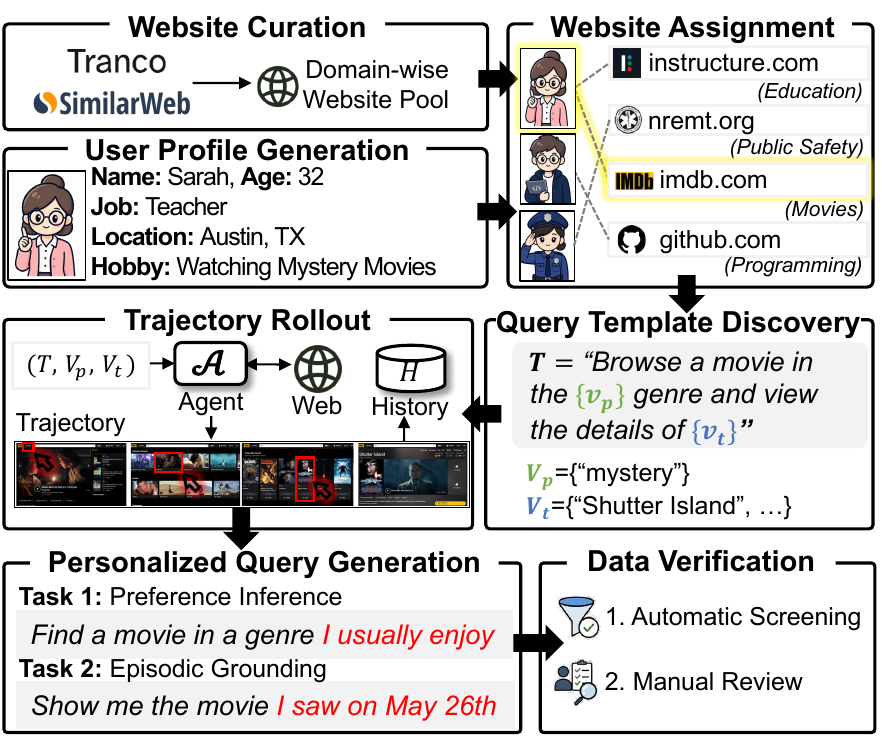}
    \caption{\benchmark construction pipeline.}
    \label{fig:data_construction}
    \vspace{-0.8em}
\end{figure}

In this section, we describe the tasks and construction process of \benchmark{}.


\subsection{Task Definition}
\label{subsec:task_definition}
We define personalized web navigation as the task of reaching a target webpage that satisfies a user query $q$, conditioned on the user's browsing history $H$.
This history $H = \{(t_i, u_i, e_i, a_i)\}_{i=1}^{L}$ comprises a sequence of entries, each recording a timestamp $t_i$, URL $u_i$, interactable element $e_i$ (e.g., a link, button), and the corresponding action $a_i$ (e.g., click, type).
Under this formulation, we evaluate web agents on two complementary tasks: preference inference and episodic grounding.

In the \textbf{preference inference} task, the query $q$ is intentionally under-specified, omitting personal preferences such as preferred websites, brands, or genres.
The agent must infer these preferences from recurring behavioral patterns in $H$ and apply them during web navigation.
For example, given the query \textit{"Find me a book in the genre I usually enjoy"} the agent should consult $H$ to determine the user's preferred genre, then navigate accordingly.
In the \textbf{episodic grounding} task, the query $q$ contains temporal or event-based references to a specific past browsing session, such as \textit{"Show me the recipe I looked at last Tuesday."} The agent must identify the corresponding entry in $H$ and re-navigate to the exact webpage.
Both task types are further divided into single-hop queries, which involve a single target website, and multi-hop queries, which require sequential navigation across two or three websites to fulfill a composite intent.

\subsection{Benchmark Construction}
\label{subsec:benchmark_construction}
As shown in Figure~\ref{fig:data_construction}, the construction pipeline integrates diverse web environments and user personas to generate high-quality evaluation data.
Throughout the pipeline, we employ Qwen3.6-27B~\citep{qwenteam2026qwen36}, a highly capable open source multi-modal LLM, as the underlying model that drives all data generation and verification components.\footnote{Further details are provided in Appendix~\ref{app:benchmark_construction_details}.}

\paragraph{Website Curation.}
We collect 8,585 candidate websites from two sources: the top 10,000 globally popular domains from the Tranco list~\cite{Le_Pochat_2019}
and the top 50 websites per category from Similarweb\footnote{\url{https://www.similarweb.com/}}.
Similarweb organizes its sites under a standardized two-level taxonomy that
we adopt: top-level domains (e.g., \textit{Arts And Entertainment}), each further
partitioned into fine-grained subdomains (e.g., \textit{Books And Literature},
\textit{Music}).
We use an LLM to classify each Tranco site into the same (domain, subdomain) structure based on its URL and extracted webpage content.
We then apply filtering to remove non-English, adult, and otherwise unsuitable sites, followed by automated browser accessibility tests that verify stable navigation under extended interaction, including resilience to CAPTCHAs and login walls at depth.
The resulting curated website pool serves as the input to all subsequent stages of our construction pipeline.

\paragraph{User Profile Acquisition and Website Assignment.}
We generate user profiles $P$ by sampling diverse personas from PersonaHub~\citep{ge2024scaling} and extracting structured attributes (i.e., age, gender, occupation, location, hobbies) using an LLM.
To construct realistic browsing histories, we associate these profiles with websites they would naturally visit.
An LLM first assigns four to eight relevant subdomains based on the persona, and then selects a specific website $W$ within each subdomain that best fits the user profile.
Furthermore, to ensure natural site-to-site transitions for multi-hop evaluation, the LLM generates cross-site storylines spanning two or three websites under a unified goal (e.g., \textit{finding a recipe and ordering the ingredients}).

\paragraph{Query Template Discovery.}
For each assigned profile-website pair $(P, W)$, an LLM-based web agent explores $W$ to discover feasible query templates.
Each template is represented as a parameterized structure $\tau = (T, V_p, V_t)$, where $T$ is a natural language query string containing placeholders, $V_p$ denotes preference variables that capture attributes a user would consistently favor across browsing sessions (e.g., preferred genre, brand), and $V_t$ denotes target variables that specify a particular item to find (e.g., a specific book title).
For instance, on a digital library site, the agent might derive the template ``\textit{Browse the} \{\textit{Genre}\} \textit{catalog and view the details of} \{\textit{Book Title}\}'', where \textit{Genre} is a preference variable and \textit{Book Title} is a target variable.
We require each template to support multiple target values under the same preference. 
Given a fixed $v_p \in V_p$, the template can be instantiated with different $v_t \in V_t$ across query instances.
For multi-hop assignments, the agent uses the cross-site scenario from the previous stage as context and explores each assigned website accordingly. 
This allows the discovered query templates to form a coherent multi-step journey across websites.

\begin{figure*}[t]
    \centering
    \includegraphics[width=\textwidth, height=5.5cm]{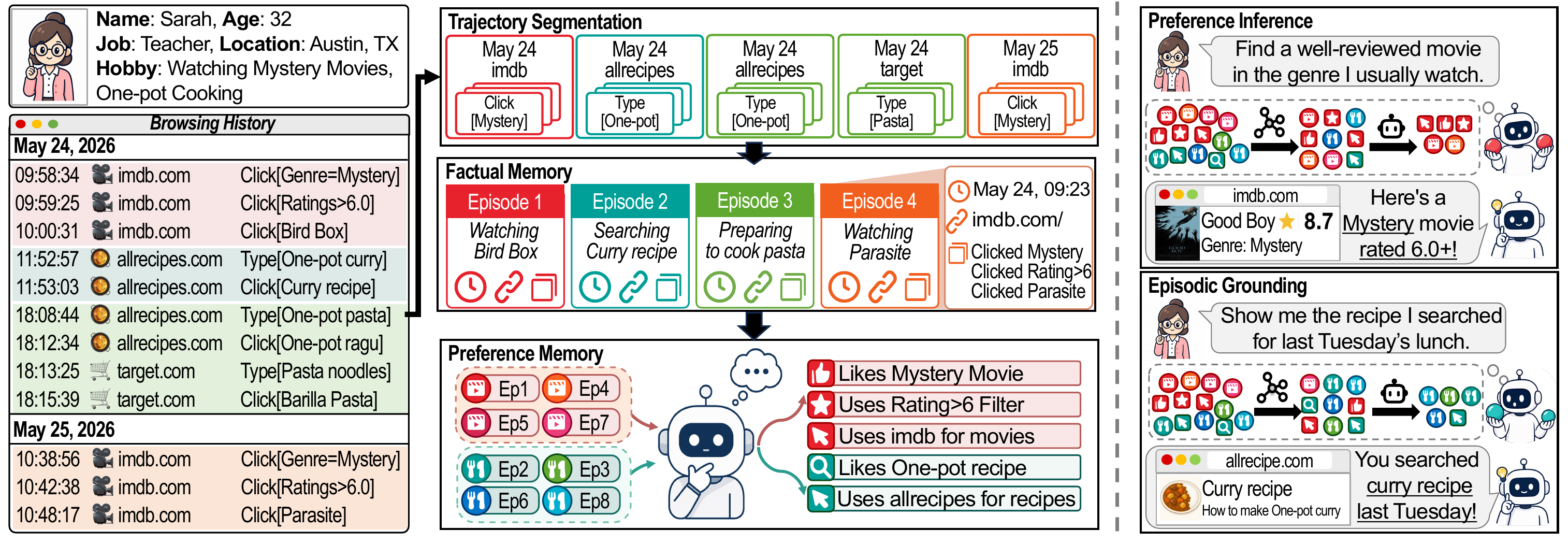}
    \vspace{-0.6cm}
    \caption{The overall architecture of \method. It first organizes raw browsing history into \emph{factual} and \emph{preference memory}, then retrieves and reranks task-relevant entries at inference time to personalize web navigation.}
    \label{fig:method_overview}
    \vspace{-0.8em}
\end{figure*}

\paragraph{Trajectory Rollout.}
For a given query template $\tau$, we instantiate its base string $T$ by substituting the placeholders with the fixed $v_p$ and a selected $v_t \in V_t$.
An LLM-based web agent $\mathcal{A}$ then executes this instantiated query on $W$.
The sequence of interactions performed during navigation is recorded as a fine-grained session $S=\mathcal{A}(W,\tau)=\{(t_i,u_i,e_i,a_i)\}_{i=1}^{\ell}$.
During execution, $\mathcal{A}$ expresses $v_p$ through site-native interaction channels (e.g., category clicks, filters), embedding the user's preference as an observable signal within session $S$.
By repeating this process for five to eight distinct $v_t \in V_t$ under the same $v_p$, we collect multiple sessions that share a consistent preference but end at varying targets.
These sessions are then annotated with simulated timestamps, where successive actions are offset by random intervals of 3 to 15 seconds to reflect realistic
pacing.
Applying this procedure across query templates assigned to $P$ yields a collection of sessions, which together constitute the user's complete browsing history $H$.

\paragraph{Personalized Query Generation.}
We use an LLM to generate natural user queries for our two personalized evaluation tasks based on the query templates $\tau$.
For the \emph{preference inference} task, we generate exactly one test query $q_{\text{pref}}$ per template $\tau$. The LLM takes the base string $T$ containing placeholders and rewrites it into an underspecified request that implicitly relies on the user's past preferences (e.g.,
\textit{``Find me a book in the genre I usually enjoy''}).
For the \emph{episodic grounding} task, the query must target a specific past interaction.
To achieve this, we sample a single $v_t \in V_t$ alongside the fixed $v_p$ to fully instantiate $T$. The LLM then rewrites this concrete description into $q_{\text{epi}}$ by incorporating temporal or event-based cues referencing the corresponding session $S$ (e.g., \textit{``Show me the book I looked at on May 25th''}).
For multi-hop assignments, both $q_{\text{pref}}$ and $q_{\text{epi}}$ are formulated as single sentences covering the entire cross-site task, following the same rules as single-hop queries.

\paragraph{Verification.}
We apply a two-stage verification process consisting of automatic screening and manual quality review. 
First, each trajectory is screened by an LLM against three conditions: (1) \emph{preference consistency} between the query, trajectory actions, and the preference values; (2) \emph{task completion}, requiring the agent to reach a terminal webpage; and (3) \emph{preference observability}, requiring $v_p$ to appear in the trajectory.
Trajectories failing any condition are discarded.
The authors then review remaining samples
against quality criteria covering preference annotations, temporal consistency in episodic grounding queries, query-trajectory alignment, and structural integrity.
Samples violating any criterion are revised or removed.
Full details of the human verification criteria are provided in Appendix~\ref{app:benchmark_verification_by_human}.

\begin{table}[t]
\centering
\small
\begin{tabular}{lrr}
\toprule
\textbf{Statistic} & \textbf{Single-hop} & \textbf{Multi-hop} \\
\midrule
Domains             & 23      & 22      \\
\raisebox{0.5ex}{$\llcorner$}{ Subdomains}          & 110     & 88      \\
Websites            & 293     & 179     \\
Users               & 100     & 100     \\
Queries             & 1{,}530 & 994     \\
Avg.\ Sessions         & 31.05   & 19.76   \\
Avg.\ Steps         & 8.23    & 19.36   \\
\bottomrule
\end{tabular}
\vspace{-0.3em}
\caption{\benchmark statistics.
\textit{Avg.\ Sessions} and \textit{Avg.\ Steps} denote the average session histories per user and interaction steps per session, respectively.
}
\label{tab:benchmark_statistics}
\vspace{-1.0em}
\end{table}

\paragraph{Benchmark Analysis.}
Table~\ref{tab:benchmark_statistics} summarizes the statistics of \benchmark{}.
Further statistical details are provided in Appendix~\ref{app:benchmark_statistics_details}.
We also compare \benchmark with real user behavior (Appendix~\ref{app:real_user_comparison}) and confirm that it requires genuine preference inference (Appendix~\ref{app:frequency_heuristics}). 

\subsection{Maintaining Benchmark Validity}
\label{subsec:benchmark_management}
A fundamental challenge in evaluating agents on the open web is reproducibility, as website changes can invalidate static annotations and render past tasks unachievable~\citep{mialon2024gaia, yangliveweb}.
To address this, we adopt a managed evaluation strategy~\citep{pan2024webcanvas} featuring a lightweight validation pipeline executed prior to each evaluation round.
First, the pipeline executes a cost-free replay to verify whether the recorded trajectories remain executable. 
For queries that fail this replay, an LLM verifies whether the underlying task can still be completed through alternative paths.
Queries that fail both checks are automatically excluded, ensuring the benchmark remains robust and reproducible.
Further details regarding this pipeline are provided in Appendix~\ref{app:benchmark_management}.

\section{\method}
\label{sec:method}


A natural approach to personalized web agent tasks is to retrieve past browsing records as external memory during web navigation~\citep{cai2025large,kim2026persona2web}.
However, our preliminary analysis reveals that even when relevant past sessions are retrieved, such memory-augmented web agents~\citep{wang2025agent,ouyang2026reasoningbank} abstract them into procedural navigation patterns, discarding preference information and episode-level details needed to fulfill personalized queries.\footnote{Error analysis of baselines is described in Appendix~\ref{app:error_analysis_of_baseline}.}
To mitigate this information loss, we introduce \method{}, which structures raw web history into personalized memory.
As illustrated in Figure~\ref{fig:method_overview}, \method operates in two distinct phases. 
The first phase, \textbf{memory construction}, distills raw browsing logs into structured factual records and behavioral preferences (\cref{subsec:memory_construction}). 
The second phase, \textbf{inference-time memory retrieval}, selects task-relevant entries from both memory types (\cref{subsec:memory_retrieval}).


\subsection{Memory Construction}
\label{subsec:memory_construction}
\method organizes user history into two distinct memory types: \textit{Factual Memory}, which captures episodic records of past actions, and \textit{Preference Memory}, which encodes recurring behavioral patterns abstracted from those actions. 

\noindent{\textbf{Trajectory Segmentation.}} Raw browsing history consists of continuous interaction logs spanning multiple websites. 
To process this effectively, we partition the continuous stream into atomic, goal-oriented segments~\citep{park2023generative,hu2025hiagent,fountas2025human}. 
Specifically, an LLM analyzes the sequence of visited webpages and executed actions to accurately identify the user's current task context. 
Whenever the model detects a semantic shift indicating a transition to a different activity, it partitions the trajectory accordingly.

\noindent{\textbf{Factual Memory Construction.}}
While individual segments provide manageable units, they often capture only fragments of a user objective.
To preserve the complete context, we group related segments sharing a high-level goal into coherent episodes.
We encode each segment with \texttt{text-embedding-3-small}~\citep{openai2024textembedding3small} and merge adjacent segments whose cosine similarity exceeds a threshold $\theta_f$.
The remaining pairs are decided by an LLM.
An LLM then turns each episode into a structured \textit{Factual Memory} entry with fields such as a descriptive title, timestamps, page URLs, and the underlying interaction sequence.
These factual records allow the agent to ground future decisions in concrete past experiences.

\noindent{\textbf{Preference Memory Construction.}}
\textit{Preference Memory} captures recurring behavioral patterns from the episodic records. 
To achieve this, we encode each factual memory entry with the same embedding model and construct a graph by treating each entry as a node.
An edge is formed between any two nodes whose pairwise cosine similarity exceeds a threshold $\theta_p$, where each connected component forms a cluster.
Finally, an LLM processes each cluster to generate a natural language memory entry that describes underlying behavioral patterns, such as a favorite platform or preferred criteria.

\subsection{Memory Retrieval}
\label{subsec:memory_retrieval}
During inference, retrieving only task-relevant memory entries is crucial to reduce noise and fit within the LLM's context window. 
To achieve this, we employ a coarse-to-fine retrieval pipeline.
First, we compute cosine similarities between the target query and the memory entries using the same embedding model from the previous stage. 
This step retrieves a broad set of top-$N$ candidates from each memory pool. 
For queries containing explicit temporal references, this candidate pool is supplemented with factual memory entries that align with temporal clues extracted from the query.
Next, an LLM evaluates these candidates, scoring them based on their semantic relevance, temporal alignment, and practical utility for the current task~\citep{sun2023chatgpt,qin2024large}.
This process yields a refined set of top-$K_f$ factual and top-$K_p$ preference memory entries to augment the agent's input context.
Detailed hyperparameter settings and analyses are provided in Appendices~\ref{app:implementation_details} and ~\ref{app:effect_of_retrieval_hyper-parameters}, respectively.

\section{Experiments} 
\label{sec:experiments}

\begin{table*}[t]
\centering
\setlength{\tabcolsep}{3.5pt}
\renewcommand{\arraystretch}{1.05}
\resizebox{\textwidth}{!}{%
\begin{tabular}{c|c|cccccccccccc}
\toprule
\multirow{3}{*}{\textbf{Models}} & \multirow{3}{*}{\textbf{Method}}
  & \multicolumn{6}{c}{\textbf{Single-hop}}
  & \multicolumn{6}{c}{\textbf{Multi-hop}} \\
\cmidrule(lr){3-8} \cmidrule(lr){9-14}
 & & \multicolumn{4}{c}{Pref. Inference} & \multicolumn{2}{c}{Epi. Grounding}
   & \multicolumn{4}{c}{Pref. Inference} & \multicolumn{2}{c}{Epi. Grounding} \\
\cmidrule(lr){3-6} \cmidrule(lr){7-8} \cmidrule(lr){9-12} \cmidrule(lr){13-14}
 & & WSR & IS & PS & TSR & ERR & TSR
   & WSR & IS & PS & TSR & ERR & TSR \\
\midrule
\multirow{4}{*}{Claude Haiku 4.5}
  & No Retrieval & 9.68 & 70.97 & 8.13 & 0.00 & 0.00 & 0.00 & 12.90 & 22.58 & 9.35 & 0.00 & 0.00 & 0.00 \\
  & AWM & 58.06 & 78.71 & 47.36 & 32.26 & 15.12 & 12.90 & 38.09 & 25.81 & \underline{58.47} & 2.50 & \underline{26.92} & \underline{21.30} \\
  & ReasoningBank & \underline{67.49} & \textbf{90.32} & \underline{62.51} & \underline{55.69} & \underline{63.07} & \underline{57.01} & \underline{52.26} & \textbf{31.94} & 58.10 & \underline{9.44} & 23.20 & 7.66 \\
  & \textbf{\method} & \textbf{74.32} & \underline{87.14} & \textbf{80.46} & \textbf{62.73} & \textbf{96.77} & \textbf{69.30} & \textbf{83.43} & \underline{29.03} & \textbf{76.23} & \textbf{19.29} & \textbf{87.10} & \textbf{41.94} \\
\midrule
\multirow{4}{*}{GPT-5.4-mini}
  & No Retrieval & 9.80 & 79.81 & 7.73 & 0.00 & 0.00 & 0.00 & 11.76 & 31.60 & 8.69 & 0.00 & 0.00 & 0.00 \\
  & AWM & 43.14 & 81.15 & 30.71 & 21.57 & 17.65 & \underline{13.73} & 19.30 & \underline{37.25} & \underline{31.80} & 0.00 & \underline{36.37} & \underline{14.80} \\
  & ReasoningBank & \underline{58.87} & \textbf{84.31} & \underline{38.05} & \underline{25.98} & \underline{38.08} & 12.60 & \underline{23.24} & 25.24 & 24.61 & \underline{1.91} & 30.37 & 4.66 \\
  & \textbf{\method} & \textbf{77.69} & \underline{82.35} & \textbf{70.22} & \textbf{58.93} & \textbf{94.12} & \textbf{71.43} & \textbf{69.16} & \textbf{50.98} & \textbf{60.07} & \textbf{26.37} & \textbf{88.24} & \textbf{23.53} \\
\midrule
\multirow{4}{*}{Gemini-3-Flash}
  & No Retrieval & 5.66 & 54.72 & 7.09 & 1.89 & 0.00 & 0.00 & 3.77 & 9.43 & 6.21 & 0.00 & 0.00 & 0.00 \\
  & AWM & 50.94 & 86.79 & 44.32 & 32.08 & 16.98 & 13.21 & 28.96 & 33.96 & \underline{50.40} & 0.73 & \underline{35.00} & 3.56 \\
  & ReasoningBank & \underline{61.79} & \textbf{98.11} & \underline{66.56} & \underline{46.88} & \underline{47.78} & \underline{37.89} & \underline{41.75} & \textbf{48.57} & 45.88 & \underline{9.19} & 31.31 & \underline{15.70} \\
  & \textbf{\method} & \textbf{79.97} & \underline{88.68} & \textbf{69.26} & \textbf{60.03} & \textbf{96.23} & \textbf{68.73} & \textbf{83.40} & \underline{43.40} & \textbf{62.75} & \textbf{22.56} & \textbf{88.68} & \textbf{52.83} \\
\midrule
\multirow{4}{*}{Gemma-3-27B}
  & No Retrieval & 7.71 & 46.80 & 6.59 & 0.78 & 0.00 & 0.00 & 9.83 & 12.47 & 9.73 & 0.00 & 0.00 & 0.00 \\
  & AWM & 42.75 & 56.08 & 27.25 & 14.25 & 23.04 & \underline{9.93} & 29.75 & 15.09 & 33.59 & \underline{0.78} & 28.74 & \underline{12.15} \\
  & ReasoningBank & \underline{64.81} & \underline{68.10} & \underline{43.12} & \underline{25.66} & \underline{32.14} & 9.92 & \underline{32.94} & \underline{16.93} & \underline{38.19} & 0.39 & \underline{31.17} & 7.29 \\
  & \textbf{\method} & \textbf{67.94} & \textbf{70.20} & \textbf{65.65} & \textbf{41.48} & \textbf{78.69} & \textbf{62.64} & \textbf{79.51} & \textbf{20.36} & \textbf{74.64} & \textbf{11.00} & \textbf{84.44} & \textbf{27.88} \\
\midrule
\multirow{4}{*}{Llama-4-Scout}
  & No Retrieval & 5.75 & 27.45 & 4.05 & 0.78 & 0.00 & 0.13 & 10.30 & 7.85 & 8.91 & 0.00 & 0.00 & 0.00 \\
  & AWM & 43.53 & 35.69 & 19.96 & 8.37 & 16.99 & 6.01 & 22.81 & \textbf{11.87} & \underline{24.77} & 0.23 & \underline{28.74} & \underline{5.31} \\
  & ReasoningBank & \underline{67.31} & \textbf{46.93} & \underline{34.35} & \underline{16.67} & \underline{36.41} & \underline{7.35} & \underline{35.01} & \underline{9.37} & 20.04 & \underline{0.39} & 27.66 & 4.55 \\
  & \textbf{\method} & \textbf{68.30} & \underline{41.70} & \textbf{42.13} & \textbf{21.60} & \textbf{76.87} & \textbf{49.83} & \textbf{72.57} & \textbf{11.87} & \textbf{45.08} & \textbf{5.56} & \textbf{85.11} & \textbf{11.07} \\
\midrule
\multirow{4}{*}{Qwen3.6-27B}
  & No Retrieval & 9.80 & 80.26 & 8.86 & 1.96 & 0.00 & 0.65 & 14.29 & 36.42 & 9.78 & 0.20 & 0.00 & 0.20 \\
  & AWM & 51.50 & 90.46 & 40.31 & 27.97 & 26.04 & 9.35 & 36.35 & \underline{43.06} & 51.25 & 1.72 & 28.74 & \underline{17.84} \\
  & ReasoningBank & \underline{67.78} & \underline{91.63} & \underline{62.49} & \underline{46.45} & \underline{42.21} & \underline{39.58} & \underline{51.39} & 42.63 & \underline{57.13} & \underline{13.73} & \underline{31.28} & 12.07 \\
  & \textbf{\method} & \textbf{71.79} & \textbf{92.42} & \textbf{64.13} & \textbf{53.61} & \textbf{80.65} & \textbf{64.84} & \textbf{84.08} & \textbf{44.67} & \textbf{76.25} & \textbf{25.41} & \textbf{87.12} & \textbf{39.03} \\
\bottomrule
\end{tabular}%
}
\caption{Main experimental results on \benchmark{}. We report Website Selection Recall (WSR), Intent Score (IS), Preference Score (PS), and Task Success Rate (TSR) for the preference inference task, and Episodic Retrieval Recall (ERR) and Task Success Rate (TSR) for the episodic grounding task, under both single-hop and multi-hop settings. \textbf{Bold} and \underline{underlined} values denote the highest and second-highest scores, respectively.}
\label{tab:main_results}
\vspace{-0.9em}
\end{table*}

In this section, we evaluate \method on \benchmark.
We first describe the experimental setup (\cref{subsec:experimental_setting}) and then report main results (\cref{subsec:main_results}).
Implementation details and prompts are provided in Appendices~\ref{app:implementation_details} and~\ref{app:prompts}, respectively.
Additional results cover a personalized memory baseline (Appendix~\ref{app:personalized_memory_baseline}), retrieval-oriented baselines (Appendix~\ref{app:retrieval_baselines}), generalization to an external benchmark (Appendix~\ref{app:external_benchmark}), error analysis of \method{} (Appendix~\ref{app:error_analysis}), and qualitative examples (Appendix~\ref{app:qualitative_results}).

\subsection{Experimental Setting}
\label{subsec:experimental_setting}
\noindent{\textbf{Baselines.}}
We evaluate \method against a \textit{No Retrieval} baseline and two memory-augmented web agent frameworks.
\textit{No Retrieval} operates without access to user history, serving as a lower bound for personalization.
For memory-augmented baselines, AWM~\citep{wang2025agent} extracts reusable workflows from past trajectories, while ReasoningBank~\citep{ouyang2026reasoningbank} distills generalizable reasoning patterns from execution traces.

\noindent{\textbf{Models.}}
We evaluate a diverse set of both proprietary and open-source backbone models.
For proprietary models, we use Claude Haiku 4.5~\citep{anthropic2025claudehaiku45}, GPT-5.4-mini~\citep{openai2026gpt54mini}, and Gemini-3-Flash~\citep{deepmind2025gemini3flash}.
For open-source models, we use Gemma-3-27B-it~\citep{gemmateam2025gemma3tr}, Llama-4-Scout~\citep{meta2025llama4}, and Qwen3.6-27B~\citep{qwenteam2026qwen36}.

\begin{figure*}[t]
    \centering
        \includegraphics[width=\textwidth]{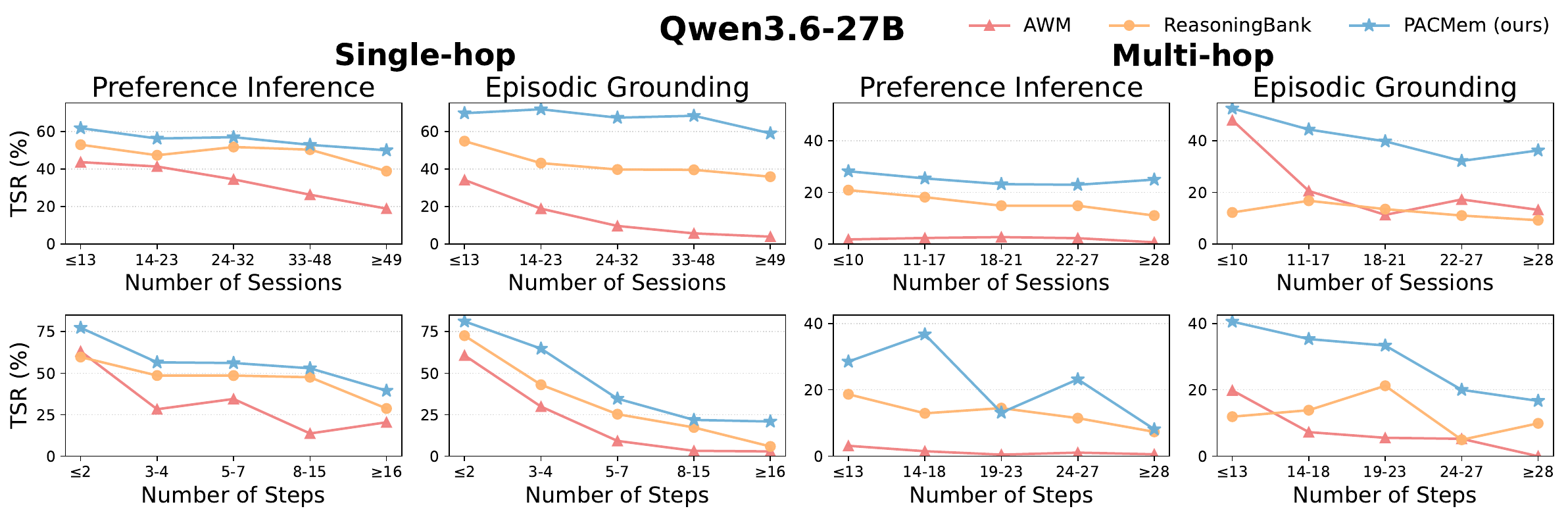}
        \vspace{-0.8cm}
        \captionsetup{skip=8pt}
        \caption{Analysis of TSR (\%) across varying numbers of sessions (top) and steps per task (bottom) on Qwen3.6-27B.
Evaluation results for Gemma-3-27B and Llama-4-Scout are provided in Figure~\ref{fig:length_anslysis_gemma} and Figure~\ref{fig:length_anslysis_llama}, respectively.}
\label{fig:length_anslysis}
    \vspace{-0.6em}
\end{figure*}

\noindent{\textbf{Evaluation.}}
We evaluate all methods using a combination of rule-based metrics and an LLM-based evaluator~\citep{zheng2023judging}, powered by Gemini 3.1 Pro~\citep{geminiteam2026gemini31pro}. 
We apply task-specific metrics to accurately evaluate both navigation and personalization performance.\footnote{Further evaluation details and LLM evaluator human-alignment are provided in Appendices~\ref{app:evaluation_details} and~\ref{app:llm-as-judge}, respectively.}

For \textbf{Preference Inference} tasks, we measure four metrics to disentangle personalization abilities from fundamental navigation success:
\begin{itemize}[leftmargin=*, nosep]
    \item \textit{Website Selection Recall (WSR):} A rule-based metric computing the recall of ground-truth target hostnames within the agent's visited websites.
    \item \textit{Intent Score (IS):} An LLM-based binary metric evaluating task completion solely from the final landing page, independent of personalization.
    \item \textit{Preference Score (PS):} An LLM-based binary metric assessing whether the agent successfully applied user preferences. It evaluates reasoning traces and the final page, explicitly crediting correct preference inference even if navigation fails.
    \item \textit{Task Success Rate (TSR):} A strict composite metric evaluating overall success, set to 1 only when all three preceding metrics are satisfied.
\end{itemize}

For \textbf{Episodic Grounding} tasks, we focus on exact memory retrieval and re-navigation:
\begin{itemize}[leftmargin=*, nosep]
    \item \textit{Episodic Retrieval Recall (ERR):} The recall score comparing the retrieved episodic entries against the ground-truth source entry.
    \item \textit{Task Success Rate (TSR):} An LLM-based binary metric verifying whether the agent successfully returned to the page referenced in the query.
\end{itemize}


\subsection{Main Results}
\label{subsec:main_results}
Table~\ref{tab:main_results} highlights a critical gap between intent satisfaction and personalization in web navigation.
The near-zero TSR across both preference inference and episodic grounding for the \textit{No Retrieval} baseline demonstrates the necessity of referencing past history.
Strong backbones like GPT-5.4-mini and Qwen3.6-27B achieve a competitive IS exceeding 79\% on single-hop queries without retrieval, yet fail to apply personal constraints.
In contrast, \method bridges the performance gap by maintaining navigation utility while improving personalization metrics such as PS and WSR.
For the preference inference task, \method{} enhances open-source models, driving Qwen3.6-27B to a 53.61\% TSR on single-hop queries and outperforming ReasoningBank, which achieved 46.45\%.
Furthermore, the episodic grounding task reveals a wide performance disparity in multi-hop scenarios, where \method{} reaches a 39.03\% TSR with Qwen3.6-27B, demonstrating grounding capabilities that exceed the 17.84\% achieved by AWM.\footnote{Multi-run variance analysis is provided in Appendix~\ref{app:variance_analysis}.}
\section{Discussion}
\label{sec:discussion}

\begin{table}[t]
  \centering
  \setlength{\tabcolsep}{4pt}
  \renewcommand{\arraystretch}{1.15}
  \resizebox{\columnwidth}{!}{%
  \begin{tabular}{cc|cc|cc}
    \toprule
    \multirow{2}{*}{\textbf{Fact.}} & \multicolumn{1}{c|}{\multirow{2}{*}{\textbf{Pref.}}}
      & \multicolumn{2}{c|}{\textbf{Gemini-3-Flash}}
      & \multicolumn{2}{c}{\textbf{Qwen3.6-27B}} \\
    \cmidrule(lr){3-4}\cmidrule(lr){5-6}
      & & Pref. Inf. & Epi. Ground. & Pref. Inf. & Epi. Ground. \\
    \midrule
    \cmark & \cmark
      & \textbf{45.27} & \textbf{62.47}
      & \textbf{42.50} & \textbf{54.67} \\[-1.6ex] 
    
    \multicolumn{6}{@{}l@{}}{%
      \makebox[0pt][l]{
        \tikz[baseline]{
          \draw[dashed] (0,0) -- (12.4cm,0); 
        }%
      }%
    }\\
    
    \cmark & \xmark
      & 37.52 (\textcolor{red}{$\downarrow$7.75}) & 56.79 (\textcolor{red}{$\downarrow$5.68})
      & 37.30 (\textcolor{red}{$\downarrow$5.20}) & 52.95 (\textcolor{red}{$\downarrow$1.72}) \\
    \xmark & \cmark
      & 36.49 (\textcolor{red}{$\downarrow$8.78}) & 28.88 (\textcolor{red}{$\downarrow$33.59})
      & 36.00 (\textcolor{red}{$\downarrow$6.50}) & 32.61 (\textcolor{red}{$\downarrow$22.06}) \\
    \xmark & \xmark
      & 29.63 (\textcolor{red}{$\downarrow$15.64}) & 10.64 (\textcolor{red}{$\downarrow$51.83})
      & 28.36 (\textcolor{red}{$\downarrow$14.14}) & 14.08 (\textcolor{red}{$\downarrow$40.59}) \\

    \bottomrule
  \end{tabular}%
  }
  \captionsetup{skip=8pt}
  \caption{Ablation study on the memory components of \method{}. We report TSR (\%), aggregated over both single-hop and multi-hop tasks. Fact. and Pref. denote factual memory and preference memory, respectively.}
  \label{tab:ablation_study}
  \vspace{-0.8em}
\end{table}

\noindent{\textbf{Impact of History and Task Complexity.}}
In real-world deployment, personalization agents must handle variations in both the volume of user history and the complexity of the target tasks.
To examine how these factors affect performance, we evaluate \method and the baselines across different numbers of session histories and numbers of steps per task.
As illustrated in Figure~\ref{fig:length_anslysis}, increasing the volume of user history universally degrades the performance.
In single-hop episodic grounding, the TSR of AWM severely drops from 34.17\% to 4.01\% as the history size increases.
A similar downward trend emerges when analyzing the number of steps per task, which serves as a proxy for task complexity.
Nevertheless, \method retains its advantage over the baselines as complexity increases.
These results suggest that distilling raw histories into structured, task-relevant memory entries remains effective under such complexity.

\noindent{\textbf{Ablation Study on Memory Components.}}
As described in \cref{sec:method}, \method refines raw browsing history into factual and preference memories.
To isolate the contribution of these memory types, we remove them and measure end-task performance.
Table~\ref{tab:ablation_study} demonstrates that combining both memory types consistently achieves the best performance. 
Excluding preference memory reduces preference inference TSR by up to 7.75\%. 
Conversely, ablating factual memory degrades preference inference moderately but causes a severe collapse in episodic grounding, decreasing by 33.59\% for Gemini-3-Flash and 22.06\% for Qwen3.6-27B. 
This shows that while preference memory captures general user patterns, factual memory remains essential for temporal and event-specific queries. 
Finally, removing both memory types leaves the web agent processing raw history directly and yields the lowest performance, validating the dual-memory design.

\section{Conclusion}
\label{sec:conclusion}
In this work, we introduce \benchmark, a benchmark for evaluating the personalization capabilities of web agents under realistic browsing conditions. 
\benchmark utilizes fine-grained browsing trajectories and evaluates agents on two key dimensions of personalization: deducing implicit user habits and recalling specific past episodes. 
Our analysis suggests that effectively managing browsing history and extracting underlying user preferences remain significant challenges. 
Alongside the benchmark, we provide \method, which constructs factual and preference memories from raw browsing histories. 
\method outperforms memory-augmented baselines, demonstrating the benefit of distilling histories into structured memory that preserves episodic details and user preferences. 
We believe that this work provides a practical foundation for developing web agents to assist individual users in real-world environments.
\clearpage

\section*{Limitations}
\label{sec:limitations}
While this work establishes a realistic benchmark for evaluating personalized web agents, several limitations remain.
First, \benchmark assumes a single, stable preference per dimension and contains only solvable tasks initiated by a single query, leaving preference drift, multiple coexisting preferences, unanswerable requests, and follow-up clarification outside the current scope.
However, \benchmark shows that current web agents struggle to infer and apply user preferences even on these static and solvable tasks, providing a foundation and future directions for personalized web agents.
Moreover, our benchmark construction pipeline can support these more complex scenarios, as histories are generated from profile specifications, simulated timestamps, and parameterized templates.
For example, preference drift can be simulated by changing a user's preference value in the middle of history generation, multiple coexisting preferences by registering more than one value for a dimension, unanswerable requests by keeping queries whose target items no longer exist, and follow-up clarification by omitting key details from the query so that the web agent must ask for them.
Building these extensions on top of the pipeline is left to future work.
Second, although the browsing trajectories are executed on live websites, the underlying user profiles and intents are synthetically generated and might not fully represent the diversity of real human populations.
Considering the privacy risks of collecting personal browsing histories, however, this synthetic construction offers a safe way to build user histories at scale.
Privacy-preserving collection of real user browsing histories remains an important direction for future work.

\section*{Ethics Statement}
Constructing the benchmark entirely on publicly accessible websites ensures the complete absence of real user data and eliminates privacy concerns.
Although reliance on large language models introduces potential demographic biases, random sampling of diverse profiles from PersonaHub effectively mitigates such risks.
Rigorous filtering mechanisms applied during the website selection phase further guarantee the exclusion of inappropriate or harmful web content.
Overall, these proactive measures ensure a safe and ethically sound evaluation environment for personalized autonomous agents.

\bibliography{custom}

\appendix

\clearpage
\startcontents[appendices]
\printcontents[appendices]{}{0}{\section*{Appendix Contents}}

\section{Implementation Details}
\label{app:implementation_details}

\paragraph{Environment Setup.}
Experiments are conducted on a server with four NVIDIA H200 GPUs running CUDA 12.9, with the agent pipeline implemented in Python 3.11.15.
We build all agents on top of Browser-Use (v0.12.6)~\citep{muller2024browseruse}, an open-source framework that exposes a unified action space for LLM-driven browser control.
To interact with the live web environment, we use Playwright (v1.59.0) as the browser automation library\footnote{\url{https://playwright.dev/}}.
We cap the maximum number of interactions at 30 steps per task to bound the navigation budget.
For inference on open-source models, we use vLLM (v0.19.0), while proprietary models are accessed through their respective APIs.

\paragraph{Hyper-parameters.}
We use \texttt{text-embedding\allowbreak-\allowbreak 3-small} as the embedding model for all retrieval operations.
For memory construction, we set the cosine similarity threshold to $\theta_f=0.75$ for factual memory grouping and $\theta_p=0.5$ for preference memory clustering.
For the retrieval pipeline, the initial dense retrieval size is set to $N=6$ candidates per memory type. 
During the subsequent LLM evaluation step, the model scores each candidate's relevance on a scale of 0 to 10, discarding any entries that fall below a minimum threshold of 5. 
From the remaining valid candidates, we select a maximum of $K_f=3$ factual and $K_p=4$ preference memory entries to augment the agent's final context.
For agent execution, we set the temperature to 0.0 across all models. 
The context window is configured to 32K tokens for Gemma-3-27B, 64K for Llama-4-Scout, 256K for Qwen3.6-27B, whereas proprietary models (Gemini, GPT, Claude) rely on default provider configurations.


\paragraph{Baselines.}
We compare \method{} against two memory-augmented baselines: AWM~\cite{wang2025agent} and ReasoningBank~\cite{ouyang2026reasoningbank}.
AWM extracts generalizable sub-routines from past action trajectories and stores them as reusable workflow templates to guide future task execution.
ReasoningBank distills strategy-level reasoning signals from both successful and failed agent experiences into structured memory items.
Both methods were originally designed to accumulate task-level procedural knowledge (reusable workflows or reasoning strategies) without any notion of user identity or personalization.
To adapt them to our personalized setting, we isolate memory pools per user profile and tag each extracted memory with episodic metadata (i.e., timestamp and URL) to support temporal queries.
For AWM, we run the workflow-induction prompt over each user's past action trajectories to build a per-user, FAISS-indexed workflow bank, and at inference time prepend the top-$k$ workflows retrieved by query embedding to the task prompt.
For ReasoningBank, we apply success and failure conditioned induction to each user's trajectories to distill structured reasoning items, and index each item by an embedding of its own induced content. 
At inference time, we retrieve and inject the top-$k$ items by cosine similarity within the user's pool.
We also include a No Retrieval baseline, in which the agent receives only the task query with no history augmentation.

\paragraph{Information Available to the Web Agent.}
At inference time, the web agent receives three inputs: the user query, the memory entries that each method derives from the browsing history (the No Retrieval baseline receives nothing from the history), and step-level browser observations.
The observations consist of the current page's screenshot annotated with Set-of-Mark~\citep{yang2023set} markers and its accessibility tree, together with the web agent's own actions taken so far.
No demographic or persona attributes are exposed to the web agent, and ground-truth labels and evaluation metadata are strictly isolated from its input.
This design reflects how personalization arises in practice.
Browsing histories accumulate naturally through everyday web usage, whereas explicit user profiles are rarely available.
Users are generally unwilling to describe their own tastes, and such self-reported profiles quickly become outdated.

\section{\benchmark Construction Details}
\label{app:benchmark_construction_details}

\paragraph{Web Agent Setup.}
All exploration and rollout agents share the same underlying configuration.
Each task is capped at 30 navigation steps to bound the agent's interaction budget, while a 60-second per-action hard timeout in the Playwright driver prevents stalls from frozen pages.
At every step, the agent receives the current page's screenshot annotated with Set-of-Mark~\citep{yang2023set} markers on all interactive elements, together with the corresponding accessibility tree listing each element's textual label and role.
This dual observation grounds the LLM's action decisions in both visual layout and structured DOM information.
To produce trajectories that resemble realistic browsing sessions, we simulate timestamps rather than recording wall-clock time: each session is anchored to a sampled start time, and successive actions are offset by 3 to 15 seconds drawn uniformly at random to mirror natural human pacing.

\paragraph{Website Assignment.}
Assigning target websites directly to user profiles from a pool of nearly 3,000 candidates is computationally expensive and prone to suboptimal matching due to the sheer volume of choices. 
To ensure accurate personalization at scale, we adopted a two-step hierarchical assignment strategy.
In the first step, the LLM maps each user profile to four to eight relevant subdomains based on high-level demographics and hobbies (e.g., a software engineer might be assigned the \textit{Consumer Electronics} and \textit{Programming} subdomains). 
In the second step, the LLM performs a fine-grained matching within these narrowed-down subdomains. 
By comparing the user's characteristics against the specific descriptions of the websites in that subdomain, the LLM selects the most suitable site. 
For instance, within the \textit{Consumer Electronics} subdomain, a profile of an audiophile is matched with a high-end audio equipment store.
This hierarchical approach guarantees that the assigned websites are not only relevant but also finely tuned to individual tastes. 
Furthermore, throughout both assignment stages, we continuously track allocation counts to ensure balanced coverage across all subdomains and websites, preventing the dataset from skewing toward a few popular domains while keeping the generation pipeline efficient.

\paragraph{Preference Consistency Across Sessions.}
A realistic browsing history for the preference inference task requires that a given user exhibit a consistent preference across all sessions on the same website.
We enforce this through a per-user preference registry: the first time the agent selects a value for $v_p \in V_p$ on a specific (user, site) pair, the value is locked into the registry, and all subsequent templates discovered or executed for the same (user, site) inherit this value rather than resampling.
Furthermore, to avoid conflicting signals, we restrict the agent from assigning multiple distinct preferences within the same semantic dimension (e.g., introducing a new movie genre once one is already established).
This guarantees that a single user's history yields a stable preference signal (e.g., always \textit{Mystery} on the digital library site), so that the preference inference task is well-posed.

\paragraph{Query Template Discovery.}
To diversify the query templates generated for a single website, the exploration agent runs for multiple iterations, with each iteration conditioned on the templates and URLs produced in previous ones.
The prompt explicitly instructs the agent to avoid duplicating prior templates and to choose a different preference dimension for each new template.
This drives the agent to explore unvisited parts of the site and surface complementary task patterns, yielding a heterogeneous template pool that collectively spans the website's distinct interaction patterns.

\paragraph{Trajectory Rollout.}
To ensure diversity across the multiple trajectories generated for the same user under a single query template $\tau$, we maintain a memory of items previously reached as $v_t$.
After each rollout, the target item at its terminal page is added to this memory, which is then injected into the prompt of the next rollout under the same template and preference $v_p$.
The agent is instructed to discover a different item satisfying $v_p$ that does not appear in the memory, thereby preventing convergence on a few popular items and yielding a heterogeneous set of trajectories that span the variety of items a real user with a consistent preference would encounter.

\begin{table}[t]
  \centering
  \small
  \setlength{\tabcolsep}{4pt}
  \begin{tabular*}{0.8\columnwidth}{@{\extracolsep{\fill}}lr@{}}
  \toprule
  \multicolumn{2}{@{}l}{\textbf{User Statistics}} \\
  \midrule
  Gender (Female / Male) & 120 / 80 \\
  Avg.\ Age              & 35.4 \\
  Min./Max.\ Age         & 14 / 67 \\
  \# Jobs                & 182 \\
  \# Locations           & 55 \\
  \# Hobbies             & 596 \\
  \midrule
  \multicolumn{2}{@{}l}{\textbf{Query \& Scenarios}} \\
  \midrule
  $|V_p|$ per user       & 8.53 \scriptsize{$\pm$ 4.79} \\
  $|V_t|$ per $V_p$      & 5.55 \scriptsize{$\pm$ 2.31} \\
  \# 2-hop scenarios     & 401 \\
  \# 3-hop scenarios     & 96 \\
  \bottomrule
  \end{tabular*}
  \vspace{-0.3em}
  \caption{Benchmark statistics of \benchmark.
  Single-hop and multi-hop tasks employ disjoint pools of 100 user profiles each, for a total of 200 distinct users.
  The $\pm$ symbol denotes standard deviation.
  $V_p$ and $V_t$ denote the preference and target variables of a query template, respectively.}
  \label{tab:dataset_stat_detail}
\end{table}

\section{Detailed Statistics of \benchmark}
\label{app:benchmark_statistics_details}
\begin{figure*}[t]
    \centering
        \includegraphics[width=0.5\textwidth]{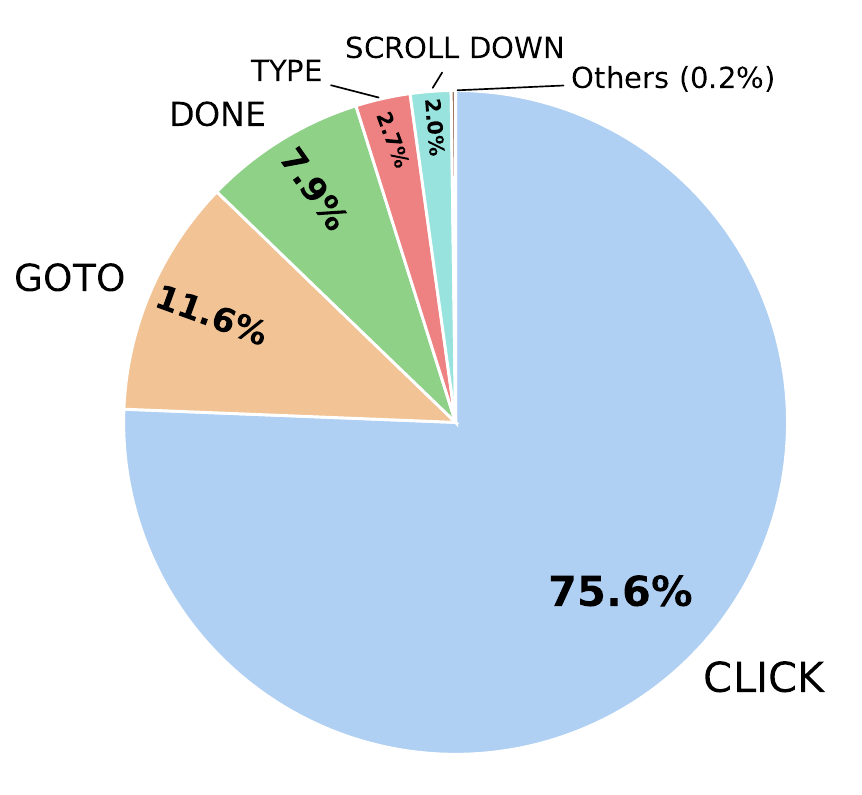}
        \caption{Action type distribution across all generated trajectories.
The \emph{Others} category aggregates PRESS, SELECT, and SCROLL UP actions, which collectively constitute 0.2\% of the total steps.}
\label{fig:action_type_distribution}
\end{figure*}

\begin{figure*}[t]
    \centering
        \includegraphics[width=\textwidth]{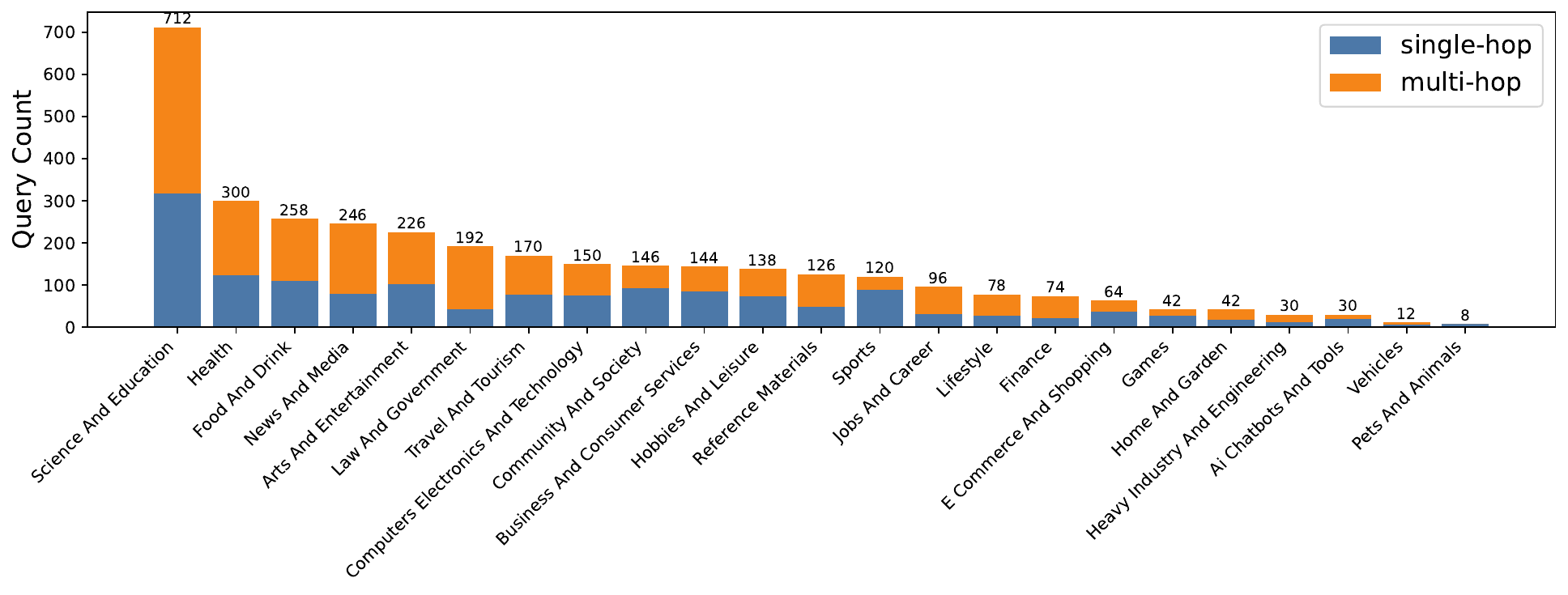}
        \caption{Number of test queries per domain in \benchmark, separated by single-hop and
   multi-hop. Multi-hop queries spanning multiple domains are counted once per distinct
  domain they touch.}
\label{fig:domain_query_distribution}
\end{figure*}

\begin{figure*}[t]
    \centering
        \includegraphics[width=\textwidth]{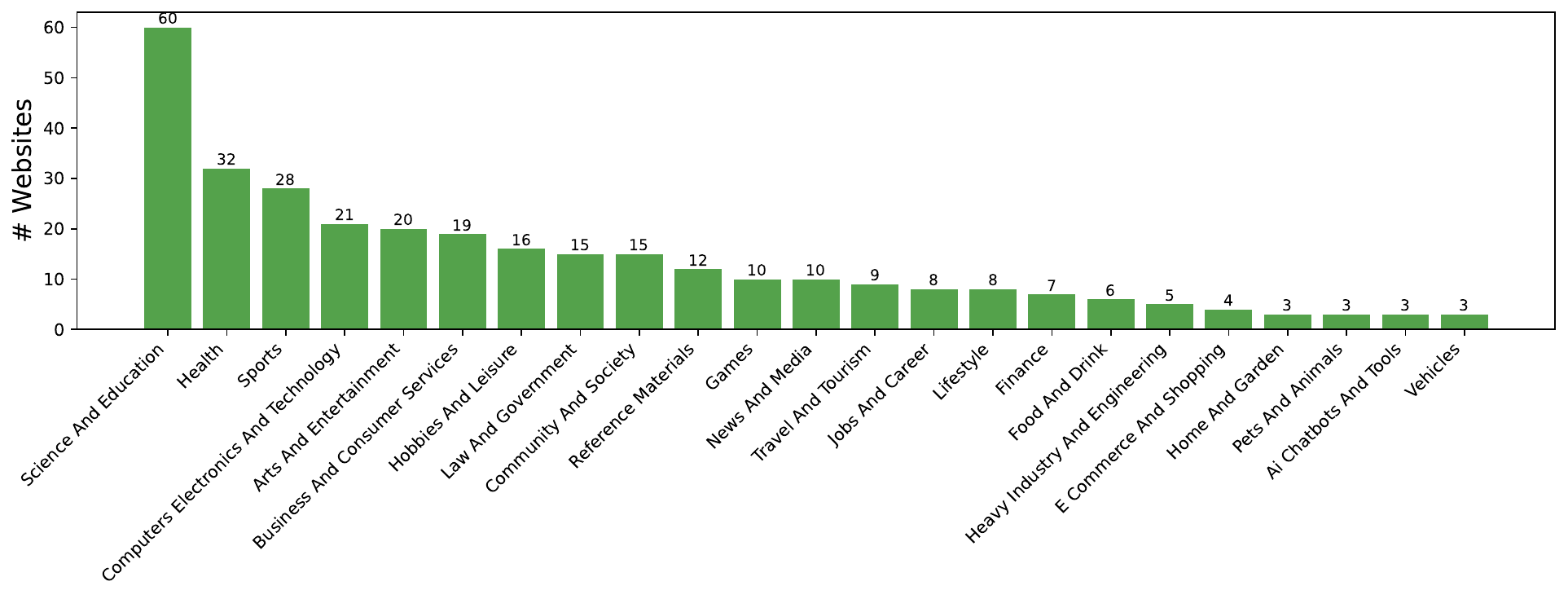}
        \caption{Number of unique websites per domain in \benchmark.}
\label{fig:domain_website_distribution}
\end{figure*}

Due to space constraints, detailed data statistics omitted from the main text are provided in Table~\ref{tab:dataset_stat_detail}.
Figure~\ref{fig:domain_query_distribution} shows the per-domain distribution of test queries, and Figure~\ref{fig:domain_website_distribution} reports the number of unique websites per domain.
Figure~\ref{fig:action_type_distribution} illustrates the distribution of action types across all generated trajectories.
Additionally, Table~\ref{tab:dataset_detail_1} details the domains, subdomains, and websites comprising the proposed benchmark.

\section{Ensuring the Realism and Validity of \benchmark}
\label{app:benchmark_verification_by_human}

\paragraph{Real-Web Grounding.}
The contents of \benchmark are grounded in real web environments.
Browsing trajectories are recordings of sessions executed on live websites, so every history entry corresponds to an interaction that actually took place, and the resulting histories naturally inherit the structure and noise of real websites.
The websites are curated from the top-ranked domains of the Tranco list and the per-category rankings of Similarweb, so the browsing environments are the widely used services that real users visit.
Test queries are rewritten from templates discovered on the actual websites, with preference and target values removed, so every query requests something that the target website supports.
As a result, an agent evaluated on \benchmark faces the same pages and interactions that it would encounter on the real web.

\paragraph{Human Verification.}
To ensure dataset reliability, we manually review all samples that pass the initial LLM filtering across four key dimensions: preference annotations validity, episodic temporal consistency, trajectory-query alignment, and structural integrity.
For preference inference tasks, we ensure preference annotation validity by checking that test queries do not leak ground-truth answers, and that trajectories reflect genuine user tastes rather than generic categories.
For episodic tasks, we verify temporal consistency, requiring that any temporal clue in the query uniquely identify a single past event without explicitly stating the target value, and that the event description accurately matches the actual sequence of actions.
Regarding trajectory-query alignment, we verify that the annotated preference is actively applied during navigation and that history entries are strictly chronological.
Crucially, we ensure that ground-truth labels and evaluation metadata are strictly isolated from the agent's input to prevent any data leakage where the agent could simply read the answers.
Finally, to maintain structural integrity, we confirm that all data fields follow the predefined schemas, cross-referenced IDs point to valid existing records, and each user meets the minimum history coverage requirement.
For multi-hop scenarios, we additionally verify that the navigation flow is logical, contains no duplicate steps, and aligns consistently with the user's profile.
During this manual review, we actively discuss and resolve any ambiguous cases.
Any sample failing to meet these criteria is either manually corrected or entirely removed.

\section{Validity Maintenance Strategy for Open Web Evaluation}
\label{app:benchmark_management}
Inspired by \citet{pan2024webcanvas}, we adopt a dedicated maintenance strategy to ensure the long-term validity of \benchmark within the dynamic open web environment. 
Because live websites frequently update their UIs, product catalogs, or layouts, static annotations can become obsolete, making past tasks unachievable. 
To mitigate this without incurring prohibitive verification costs, our pipeline operates in the following systematic stages:

\paragraph{Two-Stage Verification Pipeline.}
First, for each test query, we replay the recorded trajectories of its underlying template $\tau$ on the live website. 
This step incurs minimal cost as it requires no LLM inference. 
A test query is retained if at least three replay attempts successfully reach their recorded terminal pages.
If fewer than three replays succeed, we assume the website has evolved. 
However, UI changes do not necessarily render the underlying task infeasible. 
Therefore, we invoke a capable LLM (e.g., Gemini 3.1 Pro, Qwen3.6-27B) with a fully disambiguated version of the query—explicitly providing the ground-truth target—to verify whether the task remains achievable. 
Queries that pass either the deterministic replay or the LLM verification are retained.

\paragraph{Sample Revival.}
For queries that fail both checks, we attempt to revive the salvageable samples to prevent the benchmark from severely shrinking over time. 
If a task is still conceptually valid but the specific target is no longer available, we preserve the original query template and the user's preference attributes, updating only the target value (e.g., selecting a newly available product that matches the same preference). 
However, if a sample is fundamentally broken or infeasible to fix, we do not force a revival and instead mark it for exclusion.

\begin{table}[t]
\centering
\small
\resizebox{\columnwidth}{!}{%
\begin{tabular}{lcc}
\toprule
\textbf{Statistic} & \textbf{\benchmark{}} & \textbf{Real clickstream} \\
\midrule
Events per session, median (IQR) & 9 (5 to 16) & 8 (4 to 18) \\
Sessions revisiting a seen page & 73.4\% & 71.1\% \\
\bottomrule
\end{tabular}%
}
\caption{Browsing behavior statistics of \benchmark{} compared to real-world user logs~\citep{requena2020shopper}.}
\label{tab:real_user_comparison}
\end{table}

\paragraph{Versioning and Community Reproducibility.}
To ensure fair and transparent comparisons across future research, we actively manage these updates in our open-source repository. 
We periodically execute the validation pipeline and publish timestamped versions of the benchmark (e.g., \benchmark v1.1). 
Each release integrates the successfully revived samples and provides an explicit, updated list of the permanently excluded query IDs. 
This versioning system allows researchers to reliably reproduce evaluations on a standardized, current state of the benchmark.

\section{Statistical Comparison with Real User Behavior}
\label{app:real_user_comparison}
\begin{table}[t]
\centering
\small
\resizebox{\columnwidth}{!}{%
\begin{tabular}{llcc}
\toprule
\textbf{Extractor} & \textbf{Information given} & \textbf{Single-hop} & \textbf{Multi-hop} \\
\midrule
Majority (rule-based) & none & 6.9 & 13.2 \\
Majority (rule-based) & relevant sessions & 41.4 & 39.9 \\
Majority (rule-based) & \makecell[tl]{relevant sessions \\ + preference type} & 86.0 & 77.6 \\
LLM (Qwen3.6-27B) & none & 61.9 & 67.0 \\
\bottomrule
\end{tabular}%
}
\caption{Preference extraction accuracy (\%) of a majority-rule baseline under varying oracle conditions and an LLM extractor.}
\label{tab:frequency_heuristics}
\end{table}

While every trajectory in \benchmark{} is executed and recorded on live websites, the underlying profiles and intents are synthetically specified, which raises the question of whether the resulting histories resemble natural human browsing.
To examine this, we compare our generated histories against real user behavior, using the public Coveo clickstream dataset of real online shoppers~\citep{requena2020shopper}, which contains 443,652 real shopping sessions with 5.4M timestamped events.
On both datasets we measure two statistics, how many events a session contains and how often a session returns to an already-seen page.
Real traffic contains many one-click visits while every \benchmark{} session is a purposeful task session, so one-click visits are excluded from the real data.

As shown in Table~\ref{tab:real_user_comparison}, our generated sessions are close to real sessions in both respects.
They contain a similar number of events per session, and they return to previously seen pages at nearly the same rate as real users, rather than moving strictly forward as a scripted sequence would.
Over entire histories, 68.2\% of page visits in \benchmark{} are revisits, which lies within the 58 to 81\% range reported for real users~\citep{tauscher1997how, cockburn2001web, adar2008large}.
These results suggest that our generated histories exhibit the session structure and revisit behavior of natural human browsing.

\section{Can \benchmark{} Be Solved by Frequency Heuristics?}
\label{app:frequency_heuristics}
Preferences in \benchmark{} are expressed as repeated, observable interactions (\cref{subsec:benchmark_construction}), reflecting real browsing logs, where users rarely state their tastes explicitly and preferences must instead be inferred from behavioral signals such as clicks~\citep{10.1145/775047.775067, 10.1145/959258.959260}.
If test queries could be answered by retrieving the most repeated category from the history and taking a majority vote, \benchmark{} could be solved without genuinely inferring user preferences.
To verify that \benchmark{} requires genuine preference inference, we evaluate this majority strategy as a standalone preference extractor and report its accuracy in Table~\ref{tab:frequency_heuristics}.

The evaluation isolates preference extraction from web navigation: each extractor receives a test query and the user's entire browsing history as text, including timestamps, URLs, clicked element labels, and typed text, and must output the user's preference value.
A prediction is counted as correct when it matches the annotated preference after string normalization.
The \textit{majority extractor} involves no LLM and simply returns the most frequent interaction value in its input.
Since this strategy cannot know where to look or what to look for, we also evaluate oracle variants that hand it the answers cumulatively, first the set of relevant sessions, and then additionally the type of preference the query asks about.
The \textit{LLM extractor} receives the same input without any such help.

The plain \textit{majority extractor} fails on both single-hop and multi-hop queries (6.9\% and 13.2\%), since the most frequently repeated interactions in a history are clicks on interface elements such as search buttons, cookie banners, and menus, rather than expressions of taste.
Even when the relevant sessions are handed over (41.4\% and 39.9\%), most of the remaining errors still come from such interface clicks winning the vote.
The strategy succeeds only once the type of preference is additionally given (86.0\% and 77.6\%), since several preferences recur even within one website; the Figure~\ref{fig:intro} user, for example, repeatedly clicks both Mystery and Ratings\textgreater6.0 on the same movie site, and frequency alone cannot tell which of these signals a query refers to.
Meanwhile, the \textit{LLM extractor}, which receives no such help, remains far from ceiling (61.9\% and 67.0\%).
Majority voting therefore becomes trivial only after the actual inference is done: deciding which sessions are relevant, which type of preference the query targets, and which interactions reflect taste at all.
Making these decisions is exactly the preference inference \benchmark{} evaluates.
Moreover, extraction is only half of the task, since the inferred preference must still be applied during live navigation; across all backbones, 62 to 96\% of \method{}'s failures occur even though the evaluator judged the provided memory adequate (Appendix~\ref{app:error_analysis}).

\begin{table*}[t]
\centering
\small
\begin{tabular}{lcccc}
\toprule
\multirow{2}{*}{\textbf{Method}} & \multicolumn{2}{c}{\textbf{Single-hop}} & \multicolumn{2}{c}{\textbf{Multi-hop}} \\
\cmidrule(lr){2-3} \cmidrule(lr){4-5}
 & Preference Inference & Episodic Grounding & Preference Inference & Episodic Grounding \\
\midrule
\multicolumn{5}{l}{\textit{Gemini-3-Flash}} \\
\midrule
No Retrieval & 1.8 $\pm$ 0.4 & 0.0 $\pm$ 0.0 & 0.0 $\pm$ 0.0 & 0.0 $\pm$ 0.0 \\
AWM & 31.6 $\pm$ 1.2 & 13.6 $\pm$ 0.9 & 0.9 $\pm$ 0.4 & 3.9 $\pm$ 0.7 \\
ReasoningBank & 47.4 $\pm$ 1.1 & 37.2 $\pm$ 1.2 & 9.6 $\pm$ 0.8 & 15.1 $\pm$ 1.0 \\
\midrule
\method{} (ours) & \textbf{59.5 $\pm$ 1.2} & \textbf{68.1 $\pm$ 1.0} & \textbf{23.1 $\pm$ 1.0} & \textbf{52.2 $\pm$ 1.2} \\
\midrule
\multicolumn{5}{l}{\textit{Qwen3.6-27B}} \\
\midrule
No Retrieval & 2.1 $\pm$ 0.5 & 0.6 $\pm$ 0.3 & 0.2 $\pm$ 0.2 & 0.3 $\pm$ 0.2 \\
AWM & 27.5 $\pm$ 1.3 & 9.8 $\pm$ 1.0 & 1.9 $\pm$ 0.6 & 17.2 $\pm$ 1.2 \\
ReasoningBank & 46.0 $\pm$ 1.2 & 38.8 $\pm$ 1.3 & 13.5 $\pm$ 1.0 & 12.6 $\pm$ 0.9 \\
\midrule
\method{} (ours) & \textbf{54.2 $\pm$ 1.1} & \textbf{65.4 $\pm$ 0.9} & \textbf{24.9 $\pm$ 1.3} & \textbf{39.8 $\pm$ 1.4} \\
\bottomrule
\end{tabular}
\caption{Task Success Rate (\%, mean $\pm$ standard deviation) across three independent evaluation runs with the Gemini-3-Flash and Qwen3.6-27B backbones.}
\label{tab:variance_analysis}
\vspace{-0.6em}
\end{table*}

\section{Evaluation Details}
\label{app:evaluation_details}
This section provides the sampling protocol for our evaluation alongside implementation details for the evaluation metrics introduced in Section~\ref{subsec:experimental_setting}.

\paragraph{Stratified Sampling for Proprietary APIs.}
Evaluating web agents involves long-term navigation trajectories where input histories accumulate over successive steps, resulting in prohibitive API inference costs for proprietary models. 
To manage these expenses sustainably while maintaining a comprehensive evaluation, we evaluate proprietary models on a 25\% sampled subset of the dataset. 
This subset is carefully stratified to ensure that every user profile is represented in the evaluation.

\paragraph{Website Selection Recall.}
We extract the hostname from each visited URL by stripping the \texttt{www.} prefix and converting to lowercase, yielding a set of normalized visited hostnames.
We then compute recall as the fraction of ground-truth hostnames that appear in this set.
For single-hop queries with a single ground-truth hostname, this reduces to a binary score; for multi-hop queries, it equals the fraction of ground-truth hostnames covered by the agent's visits.

\paragraph{Intent Score.}
We employ an LLM evaluator to assign a binary score in $\{0, 1\}$ indicating whether the web agent reached an appropriate page for the given task, independently of personalization. The evaluator receives the final URL, the page text, and the task query with preference conditions removed. The score is set to 1 if the agent lands on a page that is relevant to the task intent, regardless of whether user-specific preferences were satisfied. For example, if a user queries for a restaurant reservation and the web agent lands on a relevant booking page, it receives a score of 1.

\paragraph{Preference Score.}
We employ an LLM evaluator to assign a binary score in $\{0, 1\}$ indicating whether the web agent correctly applied user preferences. The evaluator receives the user preferences, the final URL, web page information, and the full reasoning trace of the web agent formatted as a sequence of \texttt{(action, thought)} pairs. The score is set to 1 if either the final page or the reasoning trace reflects the correct preference application. For example, if a user prefers a specific brand and the web agent identifies and considers this preference during reasoning, it receives a score of 1 regardless of whether it successfully reaches the task-relevant page.

\paragraph{Episodic Retrieval Recall.}
Each episodic grounding test query is associated with a single ground-truth source entry, identified by its trajectory ID. We compute Recall between the set of trajectory IDs retrieved by the agent and the singleton ground-truth set.

\paragraph{Task Success Rate.}
For the preference inference task, TSR is set to 1 only when all three metrics (Website Selection Recall, Intent Score, and Preference Score) are exactly 1, and 0 otherwise.
For the episodic grounding task, TSR is determined by an LLM evaluator, which receives the task query, the ground-truth history entry (i.e., task description and URL), the final URL, and the page text, and verifies whether the agent successfully returned to the specific past page.

\begin{table*}[t]
\centering
\setlength{\tabcolsep}{3.5pt}
\renewcommand{\arraystretch}{1.05}
\resizebox{\textwidth}{!}{%
\begin{tabular}{l|c|cccccccccccc}
\toprule
\multirow{3}{*}{\textbf{Models}} & \multirow{3}{*}{\textbf{Method}}
  & \multicolumn{6}{c}{\textbf{Single-hop}}
  & \multicolumn{6}{c}{\textbf{Multi-hop}} \\
\cmidrule(lr){3-8} \cmidrule(lr){9-14}
 & & \multicolumn{4}{c}{Pref. Inference} & \multicolumn{2}{c}{Epi. Grounding}
   & \multicolumn{4}{c}{Pref. Inference} & \multicolumn{2}{c}{Epi. Grounding} \\
\cmidrule(lr){3-6} \cmidrule(lr){7-8} \cmidrule(lr){9-12} \cmidrule(lr){13-14}
 & & WSR & IS & PS & TSR & ERR & TSR
   & WSR & IS & PS & TSR & ERR & TSR \\
\midrule
\multirow{2}{*}{Gemini-3-Flash}
  & MemoryOS         & 43.40 & 83.02 & \textbf{71.68} & 41.51 &  64.25 & 54.72 & 45.19 & 28.85 & \textbf{70.73} &  9.62 &  22.95 & 13.21 \\
  & \textbf{\method} & \textbf{79.97} & \textbf{88.68} & 69.26 & \textbf{60.03} & \textbf{96.23} & \textbf{68.73} & \textbf{83.40} & \textbf{43.40} & 62.75 & \textbf{22.56} & \textbf{88.68} & \textbf{52.83} \\
\midrule
\multirow{2}{*}{Qwen3.6-27B}
  & MemoryOS         &  41.69 &  84.27 &  62.85 &  38.12 &  57.87 &  50.71 &  44.19 &  32.18 &  75.12 &  14.46 &  15.21 &  9.66 \\
  & \textbf{\method} & \textbf{71.79} & \textbf{92.42} & \textbf{64.13} & \textbf{53.61} & \textbf{80.65} & \textbf{64.84} & \textbf{84.08} & \textbf{44.67} & \textbf{76.25} & \textbf{25.41} & \textbf{87.12} & \textbf{39.03} \\
\bottomrule
\end{tabular}%
}
\caption{Performance comparison between \method{} and the MemoryOS baseline.}
\label{tab:memoryos_results}
\end{table*}

\section{LLM-as-Judge Alignment with Human Evaluators}
\label{app:llm-as-judge}

To assess the reliability of the LLM-based components of our evaluation, we measure both the agreement among human annotators and the agreement between the LLM judge and human annotators on each LLM-judged metric, following the protocol of \citet{zheng2023judging} and \citet{he2024webvoyager}.
For each of the three LLM-judged metrics, \textit{Intent Score} (IS), \textit{Preference Score} (PS), and the episodic grounding \textit{Task Success Rate} (TSR), we randomly sample 50 instances on which the LLM judge returns a positive verdict and 50 on which it returns a negative verdict, yielding 100 instances per metric.
Three of the authors label the instances independently, applying the same rubric and receiving the same inputs as the LLM judge (Appendix~\ref{app:evaluation_details}).
Within each metric, a randomly selected subset of 60 instances is labeled by all three annotators, and the remaining 40 instances are split across them.
We report two statistics.
First, inter-annotator reliability is measured with Fleiss' kappa ($\kappa$) on the 60 instances labeled by all three annotators, yielding $\kappa=0.87$ for IS, $\kappa=0.82$ for PS, and $\kappa=0.91$ for TSR, corresponding to almost perfect agreement~\citep{Landis1977TheMO}.
Second, we measure the judge's agreement with the human labels over all 100 instances per metric, using the majority vote on those 60 instances and the single human label on the remaining 40.
The judge agrees with the human labels on 95\% of instances for IS, 88\% for PS, and 97\% for TSR, comparable to or exceeding the agreement levels reported in recent LLM-as-judge studies for web agent evaluation~\citep{he2024webvoyager,xueillusion}.
These results indicate that the LLM judge produces verdicts closely aligned with human evaluators across both navigation completion and personalization dimensions, supporting its use as the automatic evaluator in our framework.

\begin{figure}[t]
    \centering
        \includegraphics[width=\columnwidth]{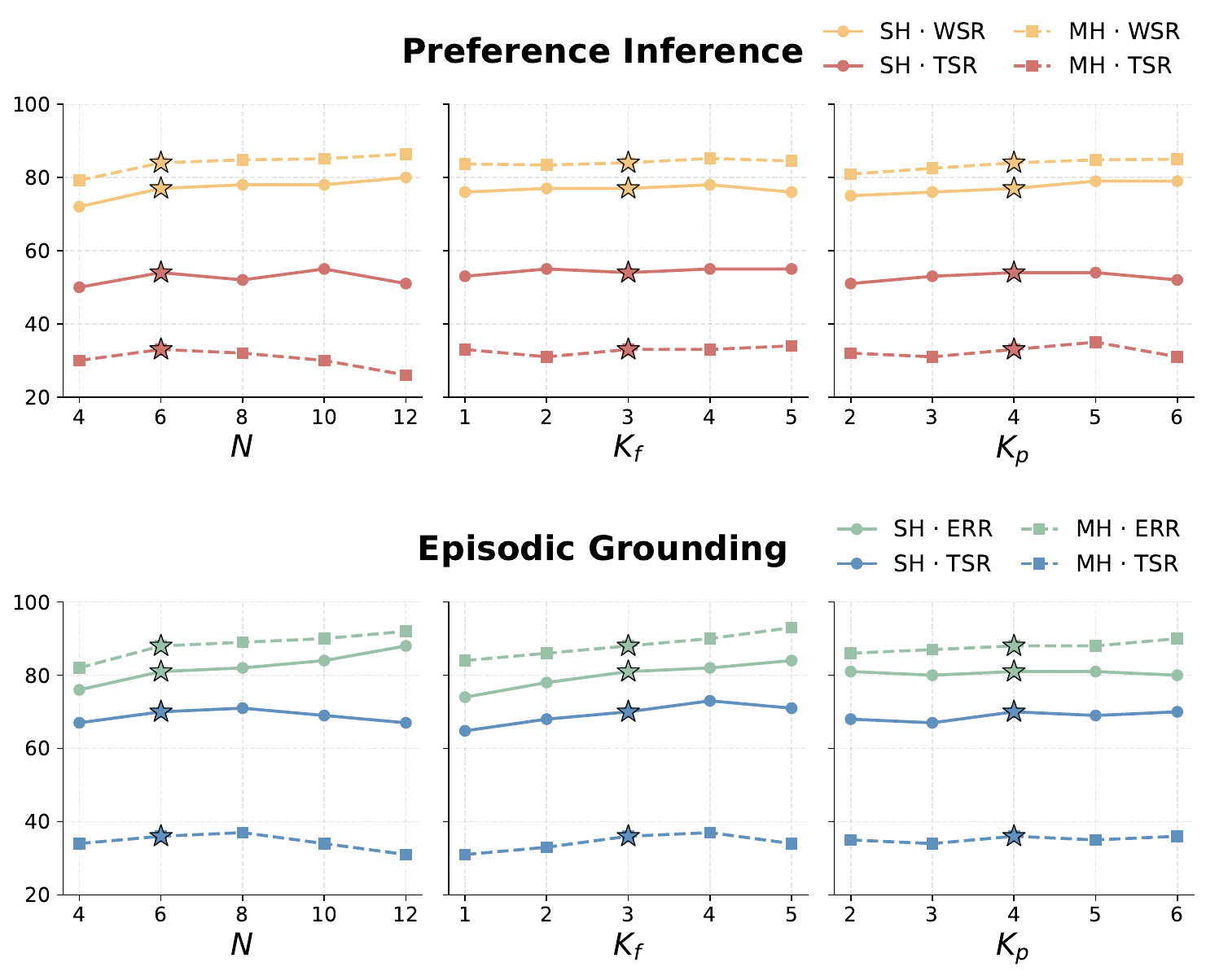}
    \caption{Performance variations of the \method across different hyper-parameter configurations.
The analysis evaluates the initial candidate size $N$, and the final selection caps $K_f$ and $K_p$ for factual and preference memories, respectively.}
    \label{fig:hyper_parameter_analysis}
    \vspace{-0.8em}
\end{figure}

\section{Multi-Run Variance Analysis}
\label{app:variance_analysis}
Since \benchmark{} is evaluated on live websites, results can vary across runs even under deterministic decoding.
To quantify this run-to-run variance, we repeat the evaluation three times and report the mean and standard deviation in Table~\ref{tab:variance_analysis}.
A full multi-run replication over every backbone is prohibitive, as a single evaluation round for one backbone and one method covers up to 2,524 live-web queries with up to 30 browser steps each, and the full grid spans six backbones and four methods.
Repeating this entire grid three times would require millions of live browser steps, each an LLM call over a screenshot, an accessibility tree, and the accumulated action history, incurring a correspondingly large token cost.
We therefore focus on Gemini-3-Flash, one of the proprietary backbones, and Qwen3.6-27B, the strongest open-source backbone in our experiments, covering four methods.
All models are decoded at temperature 0, so the variability mainly reflects live-web dynamics.

Across both backbones and all four settings, the run-to-run standard deviation stays within 1.4 points.
Among these settings, the smallest TSR difference between \method{} and a baseline is 8.2 points (54.2 vs.\ 46.0 for ReasoningBank on single-hop preference inference with Qwen3.6-27B), more than five times the largest standard deviation of 1.4 points.
Even the smallest such TSR margin across the full grid of our main experiments (Table~\ref{tab:main_results}), 4.93 points on single-hop preference inference with Llama-4-Scout (21.60 vs.\ 16.67 for ReasoningBank), is still more than three times the largest standard deviation.
Differences of this size are well beyond what the observed run-to-run variation can produce, which means the comparisons in Table~\ref{tab:main_results} remain valid across runs.

\begin{table*}[t]
\centering
\small
\begin{tabular}{lcccc}
\toprule
\multirow{2}{*}{\textbf{Method}} & \multicolumn{2}{c}{\textbf{Single-hop}} & \multicolumn{2}{c}{\textbf{Multi-hop}} \\
\cmidrule(lr){2-3} \cmidrule(lr){4-5}
 & Preference Inference & Episodic Grounding & Preference Inference & Episodic Grounding \\
\midrule
\multicolumn{5}{l}{\textit{Gemini-3-Flash}} \\
\midrule
No Retrieval & 1.9 & 0.0 & 0.0 & 0.0 \\
Raw-history RAG (BM25) & 48.1 & 14.3 & 8.4 & 0.0 \\
Raw-history RAG (Dense) & 51.4 & 16.4 & 7.9 & 4.9 \\
User-profile extraction & 46.3 & 10.8 & 12.3 & 2.6 \\
Oracle raw sessions & \textbf{72.2} & 64.6 & 20.4 & 16.8 \\
\textbf{\method{} (ours)} & 60.0 & \textbf{68.7} & \textbf{22.6} & \textbf{52.8} \\
\midrule
\multicolumn{5}{l}{\textit{Qwen3.6-27B}} \\
\midrule
No Retrieval & 2.0 & 0.7 & 0.2 & 0.2 \\
Raw-history RAG (BM25) & 41.0 & 13.7 & 14.2 & 0.0 \\
Raw-history RAG (Dense) & 46.4 & 20.8 & 10.3 & 7.4 \\
User-profile extraction & 42.1 & 16.6 & 20.2 & 1.8 \\
Oracle raw sessions & \textbf{84.5} & 61.3 & \textbf{27.1} & 15.9 \\
\textbf{\method{} (ours)} & 53.6 & \textbf{64.8} & 25.4 & \textbf{39.0} \\
\bottomrule
\end{tabular}
\caption{Task Success Rate (\%) of \method{} and retrieval-oriented baselines on \benchmark{}. No Retrieval and \method{} rows are reproduced from Table~\ref{tab:main_results} for reference.}
\label{tab:retrieval_baselines}
\vspace{-0.8em}
\end{table*}

\section{Effect of Retrieval Hyper-parameters}
\label{app:effect_of_retrieval_hyper-parameters}
\method{} incorporates specific hyper-parameters in its retrieval pipeline.
The system utilizes the initial candidate count $N$ and the final selection caps $K_f$ and $K_p$ for factual and preference memories, respectively.
We examine how varying these configurations influences performance across preference and episodic grounding tasks.
The evaluation ensures consistency by utilizing an identical pool of 100 distinct users, randomly sampling exactly one test query per user for each task.
Figure~\ref{fig:hyper_parameter_analysis} illustrates the corresponding performance variations under different parameter configurations.
Empirical results from the Qwen3.6-27B model demonstrate a clear trade-off between retrieving sufficient context and mitigating noise.
Setting the initial candidate pool ($N$) too small restricts the available memory context, decreasing recall metrics such as WSR.
Conversely, expanding the candidate pool largely introduces irrelevant interactions, which consistently degrades the TSR across both tasks.
Evaluating the memory caps reveals that increasing $K_f$ for episodic grounding and $K_p$ for preference inference initially improves performance by supplying broader historical context, but excessively large limits degrade the TSR due to noise.

\section{Comparison with a Personalized Memory Baseline}
\label{app:personalized_memory_baseline}
The main experiments evaluate \method{} against web agent baselines equipped with memory modules designed for general web tasks.
To assess personalization capabilities more rigorously, we extend the evaluation to include a baseline explicitly designed for long-term personalization.
Specifically, we benchmark against MemoryOS~\citep{kang-etal-2025-memory}, a framework that achieves personalized memory retention in conversational agents through a hierarchical storage architecture.
As illustrated in Table~\ref{tab:memoryos_results}, \method{} achieves higher TSR than MemoryOS in every setting.
While MemoryOS attains a higher Preference Score on the Gemini-3-Flash backbone, it fails to ground implicit preferences into actionable web navigation steps.
Furthermore, the pronounced performance gap observed in multi-hop episodic grounding scenarios reveals that dialogue-targeted memory representations lack the structural granularity required to reconstruct exact web interactions.
These findings validate the necessity of specialized structural memory designs for personalized web operations.

\begin{figure*}[t]
    \centering
        \includegraphics[width=\textwidth]{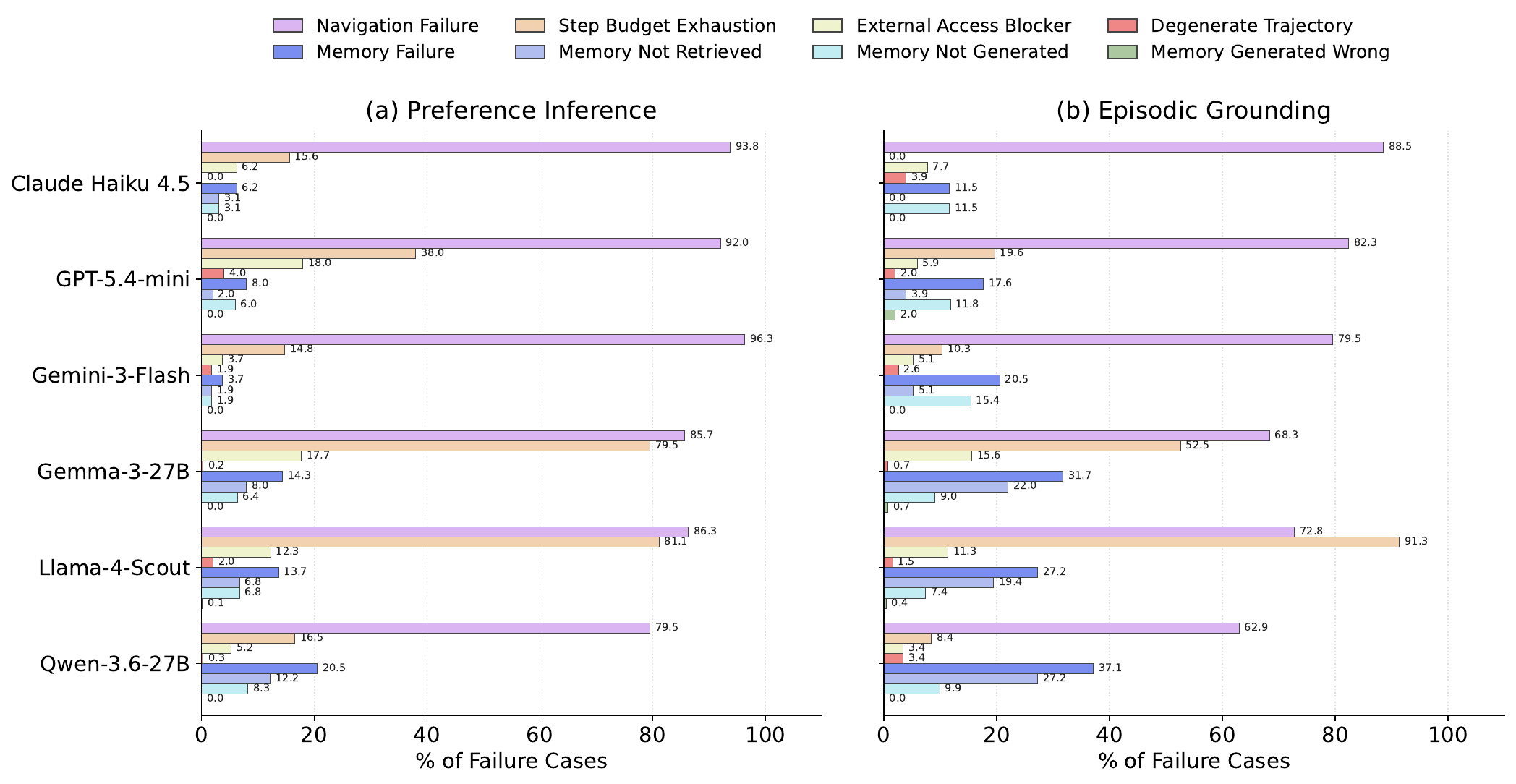}
    \caption{Failure-mode distribution of \method per backbone model on (a) preference inference and (b) episodic grounding tasks, aggregated over single-hop and multi-hop settings.}
    \label{fig:error_analysis}
\end{figure*}

\section{Experiments with Retrieval-Oriented Baselines}
\label{app:retrieval_baselines}
Beyond the memory-augmented frameworks in our main experiments, personalization from browsing history could in principle be handled by standard text-retrieval methods applied directly to the raw browsing logs.
We additionally evaluate such retrieval-oriented baselines on \benchmark{}.
Since AWM, ReasoningBank, and MemoryOS in our experiments already summarize raw trajectories into retrievable memory items and thus cover summary-based RAG, this section covers the remaining settings of raw-history RAG, user-profile extraction, and oracle raw sessions.

\textit{Raw-history RAG} retrieves relevant segments of the raw logs with either BM25 or dense retrieval using text-embedding-3-small.
\textit{User-profile extraction} prompts the backbone to read the full history once and extract a preference profile, which then augments the web agent's input context.
\textit{Oracle raw sessions} directly provide the ground-truth relevant sessions to the web agent.
We use the Gemini-3-Flash and Qwen3.6-27B backbones.
All settings run with the same web agent as in our main experiments and differ only in the content injected into its input context.

As shown in Table~\ref{tab:retrieval_baselines}, \textit{raw-history RAG} and \textit{user-profile extraction} show substantially lower performance on episodic grounding and multi-hop queries.
Notably, even \textit{oracle raw sessions} fall behind \method{} on episodic grounding despite receiving the ground-truth sessions, indicating that the bottleneck lies in how the history is represented rather than in retrieval itself.
\textit{User-profile extraction} discards individual episodes and their timestamps during abstraction, so episodic grounding collapses to 2.6\% and 1.8\% on multi-hop queries.
\textit{Raw-history RAG} preserves the original logs instead, but the web agent must re-interpret low-level events at inference time, the same burden that limits the oracle setting.
Distilling the history into structured entries while retaining the timestamps and URLs of each episode avoids both failure sources.

\section{Generalization of \method{} to an External Benchmark}
\label{app:external_benchmark}
\begin{table}[t]
\centering
\small
\resizebox{\columnwidth}{!}{%
\begin{tabular}{llccc}
\toprule
\textbf{Model} & \textbf{Memory} & \textbf{\pavg} & \textbf{Intent} & \textbf{SR} \\
\midrule
\multirow{3}{*}{Gemini-3-Flash} & no-history & 17.5 & 46.3 & 0.0 \\
 & pre-execution & 65.3 & 43.1 & 10.8 \\
 & \textbf{\method{} (ours)} & \textbf{74.1} & \textbf{53.0} & \textbf{16.2} \\
\midrule
\multirow{3}{*}{Qwen3.6-27B} & no-history & 11.4 & 31.4 & 0.0 \\
 & pre-execution & 59.7 & 35.0 & 7.1 \\
 & \textbf{\method{} (ours)} & \textbf{68.2} & \textbf{46.7} & \textbf{13.3} \\
\bottomrule
\end{tabular}%
}
\caption{Performance (\%) of \method{} and baseline history access schemes on the Persona2Web benchmark.}
\label{tab:external_benchmark}
\end{table}
We additionally evaluate \method{} on Persona2Web~\citep{kim2026persona2web} to examine whether its memory design remains effective beyond \benchmark{}.
Persona2Web is an external benchmark that evaluates personalized web agents on the open web with LLM-synthesized user histories of predefined action types.
Since its records are discrete natural-language events rather than continuous interaction logs, the only adaptation is at input formatting.
We group the records into daily sessions, after which memory construction and retrieval run unchanged, and the retrieved memories augment the web agent's input context as in our main experiments.
As baselines, we adopt \textit{pre-execution} and a \textit{no-history} lower bound.
\textit{Pre-execution}, one of the history access schemes proposed in the original paper, retrieves relevant history entries by generating retrieval queries before execution begins.
Following the original evaluation protocol, we report the average personalization score (\pavg), computed as the mean of the website and preference personalization scores, the intent satisfaction score (Intent), which measures task completion independently of personalization, and the success rate (SR), which counts a task as successful only when both personalization and intent are fully satisfied.

As shown in Table~\ref{tab:external_benchmark}, \method{} outperforms \textit{pre-execution} on all three metrics for both backbones.
Since these gains arise on histories generated by an entirely different pipeline in a different format, they indicate that the memory design remains effective beyond \benchmark{}.

\section{Error Analysis of \method}
\label{app:error_analysis}
To investigate the failure modes of \method{}, we analyze incorrect prediction samples and categorize them using an LLM (i.e., Gemini 3.1 Pro).
Specifically, we evaluate failures (i.e., TSR\,=\,0) across all six backbone models, covering both task types in single-hop and multi-hop settings.
Each failure is classified along two complementary dimensions.
First, three script-level checks inspect the agent trajectory directly: \textit{Step Budget Exhaustion} (the agent hits the step limit due to repetitive or looping behavior), \textit{External Access Blocker} (the agent is intercepted by CAPTCHA, Cloudflare, or similar bot defenses), and \textit{Degenerate Trajectory} (the agent never performs meaningful navigation).
Second, the LLM evaluator is applied in two stages.
Stage 1 receives the user query, the retrieved factual and preference memories, the ground-truth target, the agent's full trajectory, and the LLM-as-Judge reasons.
It labels each failure as either a \textit{Navigation Failure} (the memory provided was adequate but the agent failed to navigate to the target) or a \textit{Memory Failure} (the memory provided was insufficient or wrong).
Stage 2 further subdivides memory failures by comparing the user's full memory pool against the actually retrieved entries: \textit{Memory Not Generated} (no relevant entry exists in the pool), \textit{Memory Not Retrieved} (a relevant entry exists but was not ranked into the top-k), and \textit{Memory Generated Wrong} (the relevant entry was retrieved but its content materially diverges from the ground truth, decided by the LLM evaluator).
\textit{Navigation Failure} and \textit{Memory Failure} partition the failure cases, and the three subcategories partition \textit{Memory Failure}.
The script-level flags can overlap with the LLM-evaluator labels and with each other, so the bars per model can sum above 100\% (e.g., a navigation failure that also exhausted the budget).
The two dimensions capture different aspects: script-level flags describe \emph{how} the failure surfaced, while the LLM-evaluator labels identify \emph{why} the agent failed.

Across all backbones, \textit{Navigation Failure} is the dominant root cause (62\% to 96\%, Figure~\ref{fig:error_analysis}), indicating that most failures arise from the navigation process itself rather than from the memory provided to the web agent.
Tables~\ref{tab:case_study_1} and~\ref{tab:case_study_2} illustrate two such navigation failures, in which the retrieved memory explicitly states the user's preference yet the web agent ends up on a page that does not match it.
An evaluation that measures only retrieval cannot capture this type of failure, since the relevant memory entries were correctly retrieved in both cases.
In contrast, our end-to-end setting evaluates whether the web agent applies the retrieved information during navigation, revealing the weaknesses of web agents and directions for improvement.
On episodic grounding, the same separation appears in Table~\ref{tab:main_results} as the gap between ERR and TSR: ERR measures only retrieval, while TSR requires the web agent to return to the referenced page.
Table~\ref{tab:case_study_4} shows a navigation failure in which the correct entry containing the exact target URL and timestamp is retrieved (ERR $=1$), yet the web agent never reaches the page (TSR $=0$), widening the gap between the two metrics.
Within memory failures, \textit{Not Retrieved} matches or exceeds \textit{Not Generated} on the open-source backbones with sufficient sample size (e.g., 22.0\% vs.\ 9.0\% for Gemma-3-27B on episodic grounding), pointing to retrieval and reranking rather than upstream memory construction as the limiting step.
Table~\ref{tab:case_study_3} illustrates such a case on episodic grounding, in which the entry for the referenced session exists in the memory pool but is not ranked into the retrieved set.
\textit{Memory Generated Wrong} is negligible across all cells ($\leq$2.0\%), confirming that when relevant entries are present they are well-formed.
The script-level flags reveal a model-capacity effect.
\textit{Step Budget Exhaustion} reaches 52 to 91\% on the weaker open-source backbones (Gemma-3-27B, Llama-4-Scout), reflecting their tendency to enter unproductive loops.
\textit{External Access Blocker} remains a uniform 3 to 18\% across all backbones, behaving as an environmental constant rather than a model-specific weakness.

\section{Error Analysis of Baseline}
\label{app:error_analysis_of_baseline}
\begin{figure*}[t]
    \centering
        \includegraphics[width=0.6\textwidth]{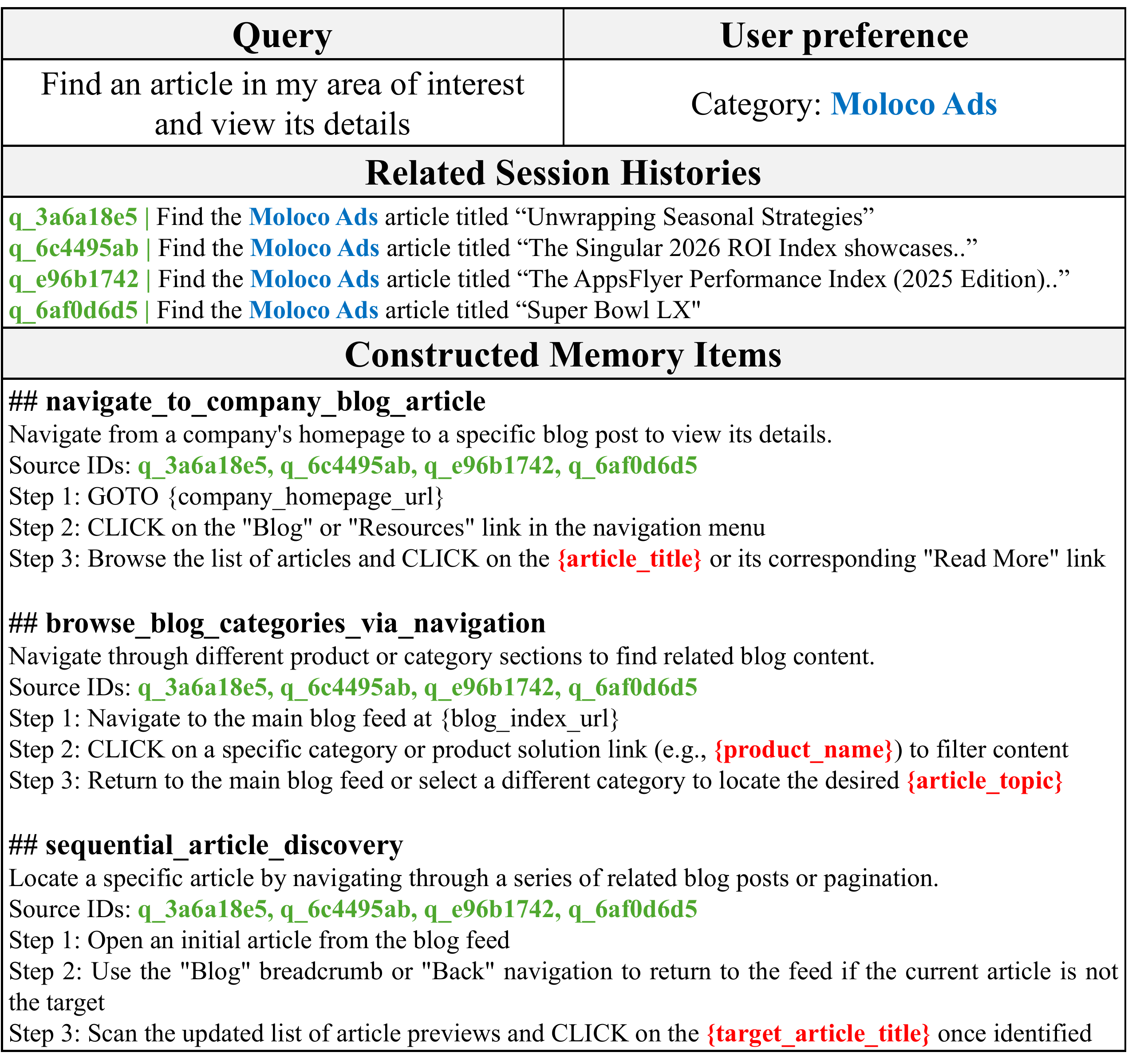}
        \caption{A representative failure case of AWM on a preference inference query. Blue marks the user's preference. Green marks the query IDs of past sessions sharing this preference, whose natural-language descriptions are synthesized into trajectories from which AWM constructs its workflow memories. Red marks the unfilled placeholder slots in the resulting memory items.}
\label{fig:pref_awm}
\end{figure*}

\begin{figure*}[t]
    \centering
        \includegraphics[width=0.6\textwidth]{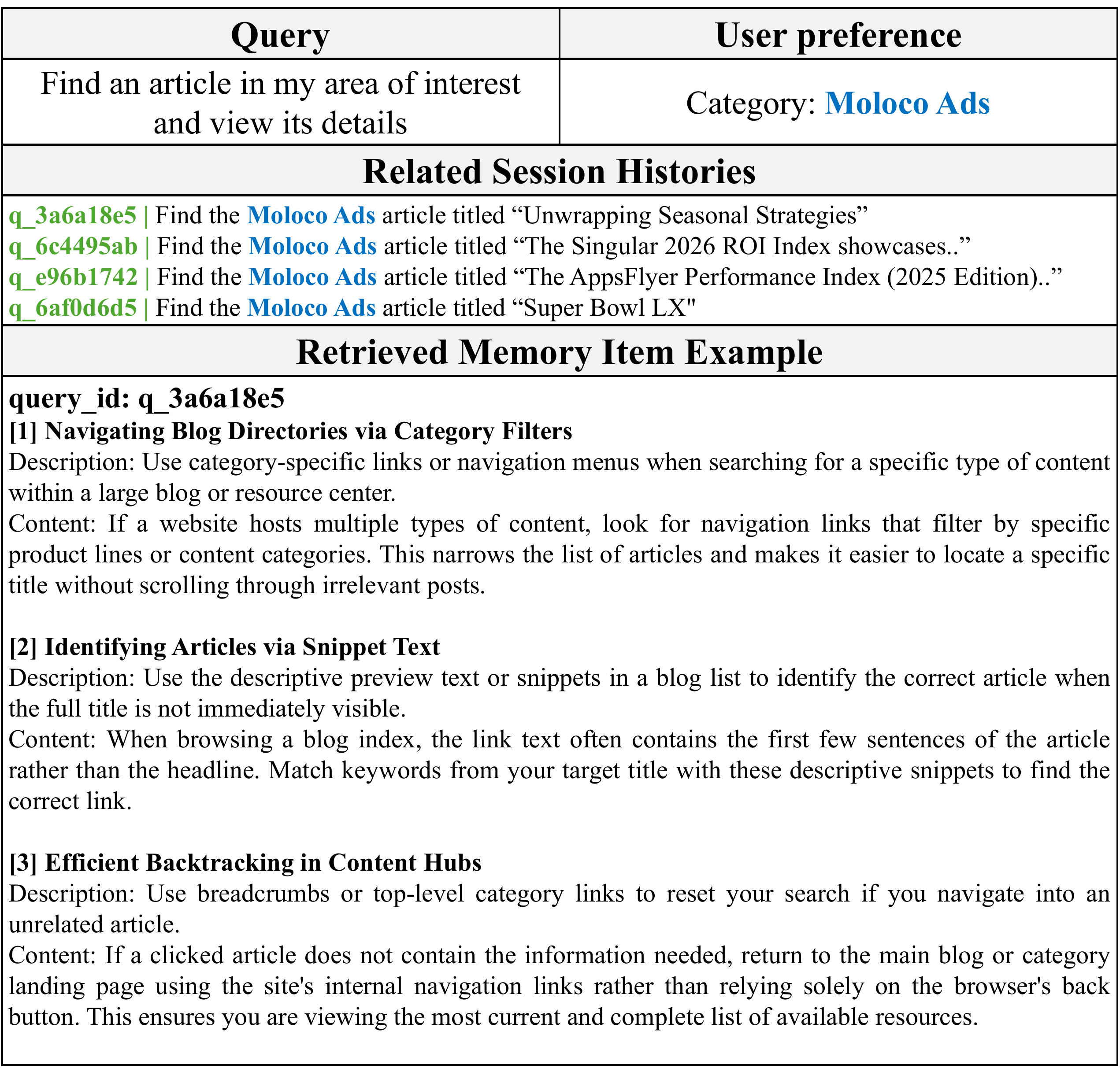}
        \caption{A representative failure case of ReasoningBank on a preference inference query. Blue marks the user's preference. Green marks the query IDs of past sessions sharing this preference, whose natural-language descriptions are synthesized into trajectories from which ReasoningBank distills its memory items.}
\label{fig:pref_rb}
\end{figure*}

To better understand the behavior of existing memory-based web agents on personalized tasks, we analyze failure cases by running inference on \benchmark with AWM and ReasoningBank.
\paragraph{Preference Inference Query.} Figures \ref{fig:pref_awm} and \ref{fig:pref_rb} illustrate representative failures of AWM and ReasoningBank on a preference inference query, respectively.
The test query, \textit{"Find an article in my area of interest and view its details"}, contains no content-specific keyword, so the agent must recover the user's preference (blue) from the web interaction history.
The user's history contains four sessions sharing this preference (green); their natural-language descriptions are synthesized into trajectories that both AWM and ReasoningBank ingest when constructing memory.

AWM pools these trajectories into workflow memories describing reusable navigation patterns such as locating the blog link, clicking a category filter, backtracking via breadcrumbs.
Each workflow correctly references source IDs, but the preference bearing attributes are reduced to unfilled placeholder slots (red).
The workflows retain how to browse the blog but not what category the user prefers.

ReasoningBank exhibits the same failure under a different memory organization. Each past session becomes one entry with (Title, Description, Content) lessons, and the retrieved lessons describe only the procedure.
The source session mentions \textit{"Moloco Ads"}, but the lesson text refers instead to \textit{"specific product lines"}, \textit{"your target title"}, or \textit{"the correct article"}.
These referring expressions appear concrete yet bind to no actual value, hiding the missing grounding beneath natural language rather than exposing it as an empty slot.

In both methods, the preference is never written into memory.
Even with complete source-ID coverage (AWM) or correct retrieval of the relevant entries (ReasoningBank), the agent receives only navigation procedure with no signal that \textit{"Moloco Ads"} is the category to filter for, and the preference cannot be applied.

\begin{figure*}[t]
    \centering
        \includegraphics[width=0.55\textwidth]{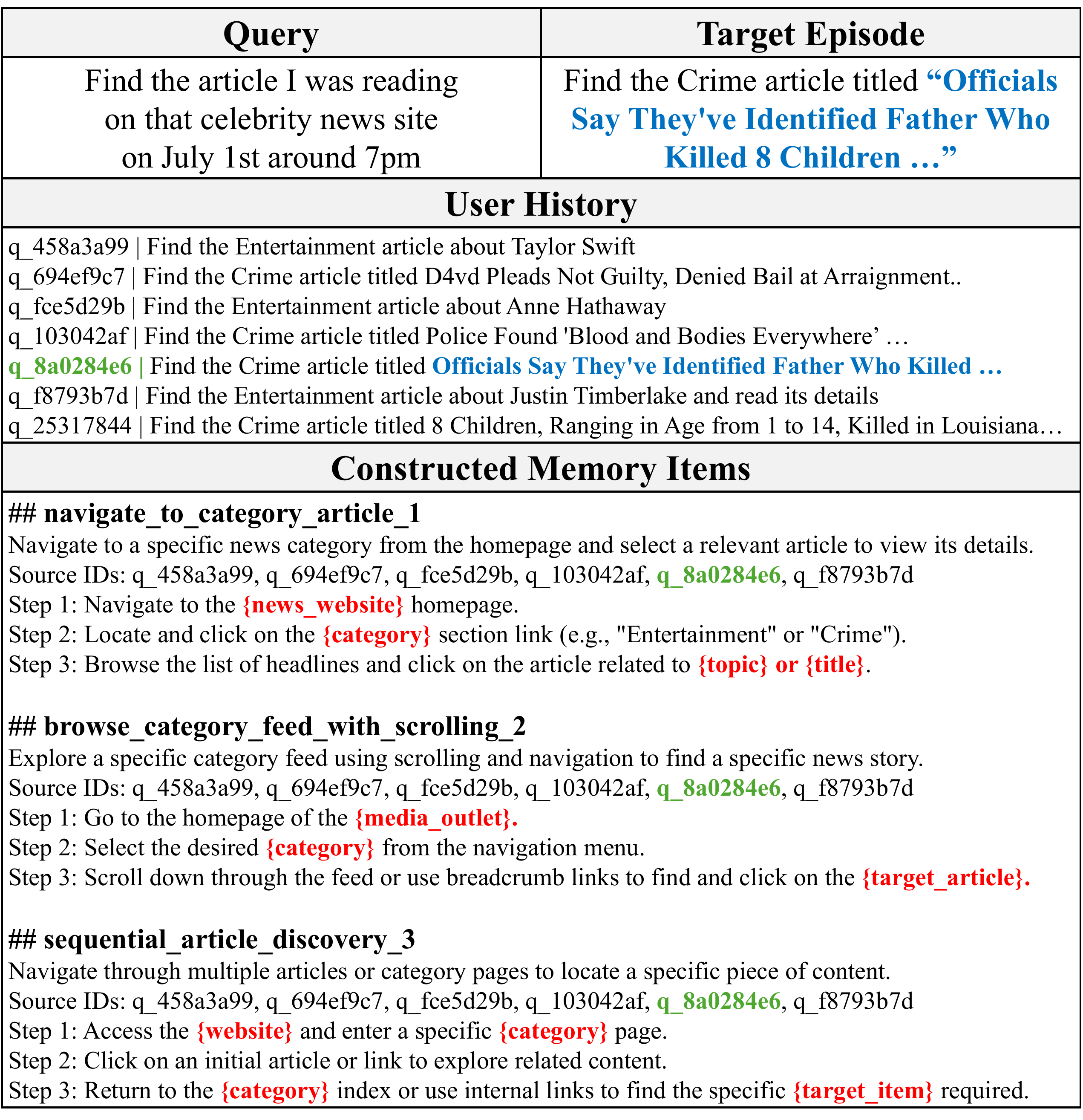}
        \caption{A representative failure case of AWM on an episodic-grounding query. Blue marks the target episode the agent must recall. Green marks the query ID of the same episode in the user history, whose natural-language description is synthesized into a trajectory from which AWM constructs its workflow memories. Red marks the unfilled placeholder slots in the resulting memory items.}
\label{fig:ep_awm}
\end{figure*}

\begin{figure*}[t]
    \centering
        \includegraphics[width=0.55\textwidth]{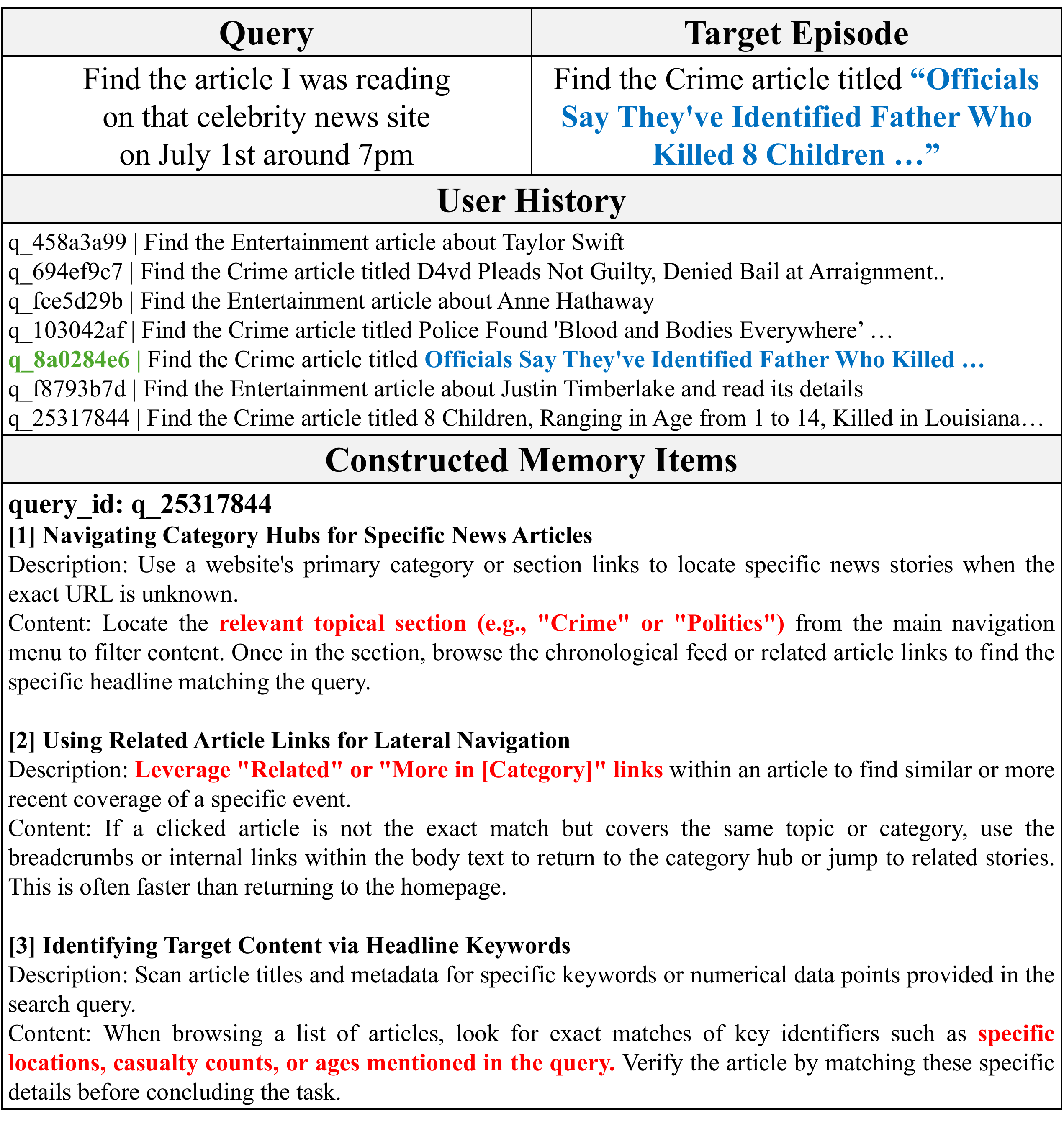}
        \caption{A representative failure case of ReasoningBank on an episodic-grounding query. Blue marks the target episode the agent must recall. Green marks the query ID of the same episode in the user history, whose natural-language description is synthesized into a trajectory from which ReasoningBank distills its memory items. Red marks referring expressions in the memory items that superficially appear to denote concrete attributes but bind to no actual value from the source episode.}
\label{fig:ep_rb}
\end{figure*}

\paragraph{Episodic Grounding Query.}
Figures \ref{fig:ep_awm} and \ref{fig:ep_rb} illustrate representative failures of AWM and ReasoningBank on an episodic grounding query, respectively.
The test query identifies the target episode by cues about when and what the user previously visited rather than by direct content, so the agent must recall a specific past article (target episode; blue).
The user's history contains seven prior sessions; one of them is the target episode (green), and its natural-language description is synthesized into a trajectory that both AWM and ReasoningBank ingest when constructing memory.

AWM pools the sessions into workflow memories describing reusable news-site navigation patterns such as visiting the homepage, clicking a category, scrolling the feed. Each workflow correctly references the target source ID, but the episode-specific attributes are reduced to unfilled placeholder slots(red).
The workflows retain how to navigate a news site but not which past visit matches the cues in the query.

ReasoningBank exhibits the same failure under a different memory organization.
Each past session becomes one entry with (Title, Description, Content) lessons, and the target entry is correctly retrieved at inference time.
The lessons, however, again describe only procedure. Phrases that superficially appear to denote concrete attributes(red) employ referring expressions that read as specific yet bind to no actual value from the source episode. As in the preference-inference case, the missing grounding is hidden beneath natural language rather than exposed as an empty slot.

In both methods, the cues in the query have no mapping into memory: no per-episode attribute is encoded in a form the agent can use.
Even with the target source ID correctly referenced (AWM) and the target entry correctly retrieved (ReasoningBank), the agent receives only navigation procedure and cannot recall the specific past article.

\section{Qualitative Results}
\label{app:qualitative_results}
\begin{figure*}[t]
    \centering
        \includegraphics[width=0.8\textwidth]{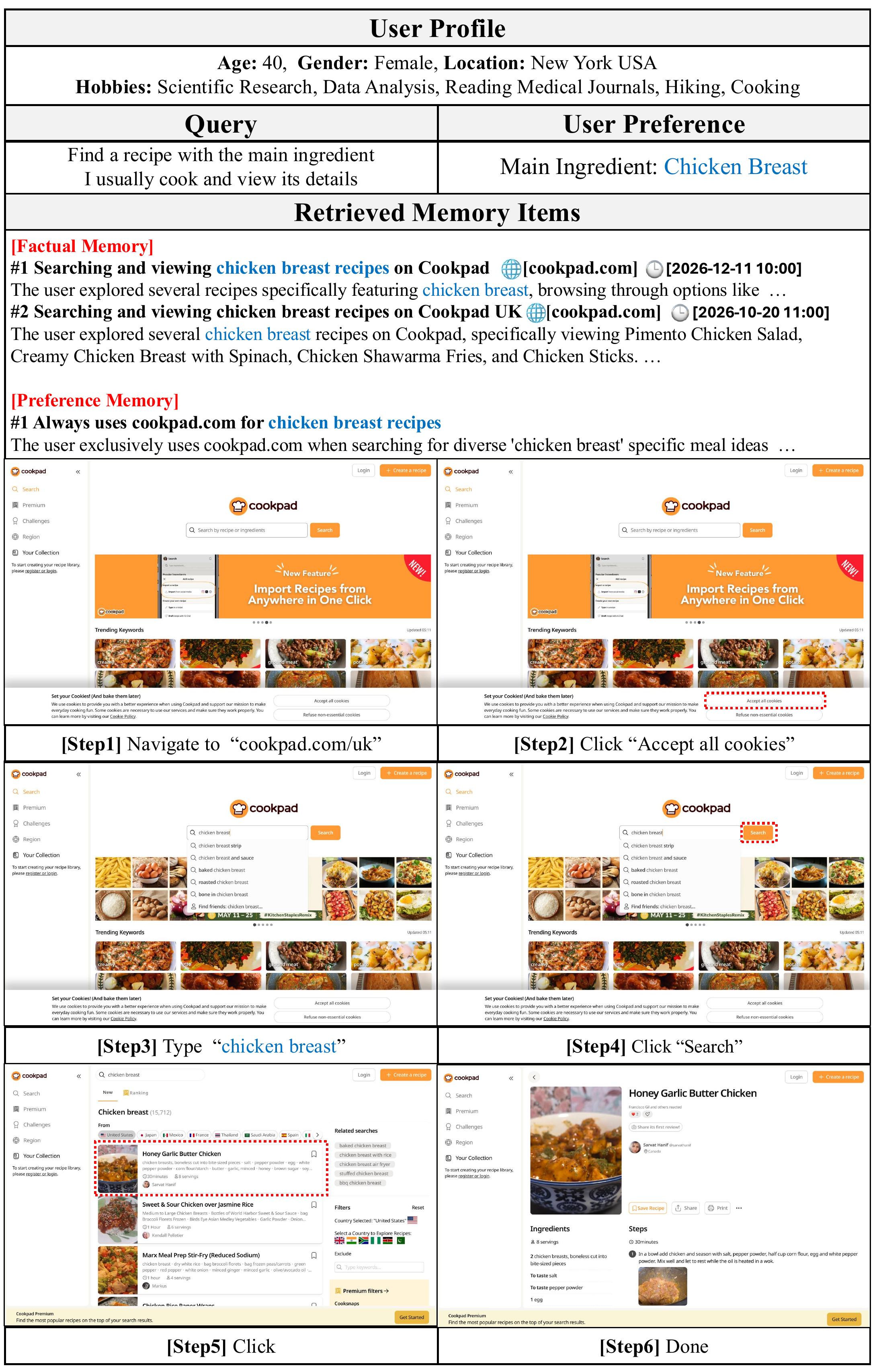}
        \caption{Execution trajectory of \method{} for a single-hop preference inference task. Given the query \textit{"Find a recipe with the main ingredient I usually cook"}, \method{} retrieves the historical preference for \texttt{chicken breast} as the main ingredient from preference memory.
The agent dismisses the cookpad.com cookie banner (Step 2), types \textit{chicken breast} into the recipe search box (Step 3), and submits the query (Step 4).
Subsequently, the agent selects a matching recipe from the results (Step 5) and lands on the target recipe page (Step 6).
For visualization purposes, target UI elements for click actions are highlighted with red dashed bounding boxes.
}
\label{fig:qualititative_sh_pref}
\end{figure*}

\begin{figure*}[t]
    \centering
        \includegraphics[width=0.73\textwidth]{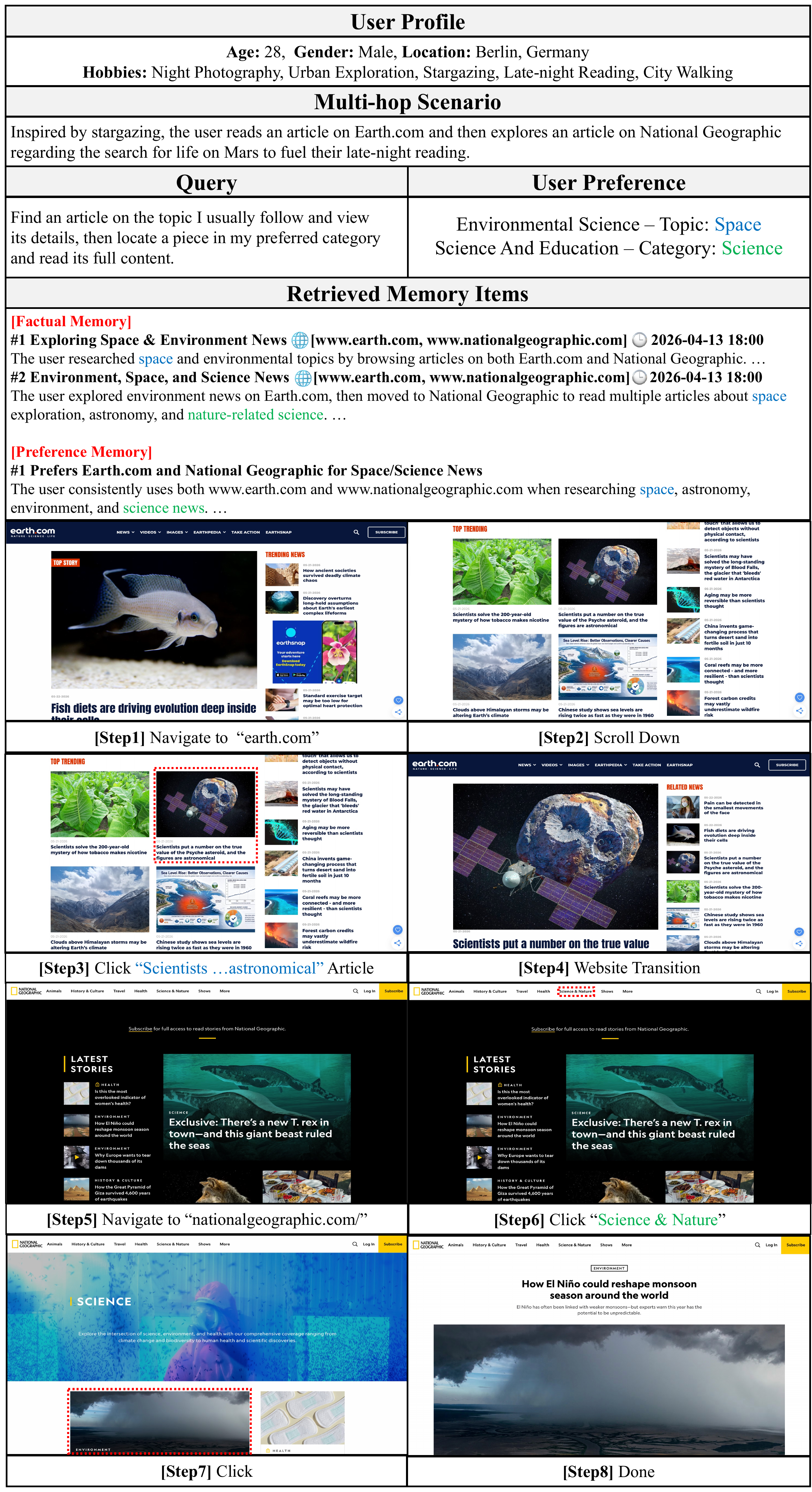}
        \caption{Execution trajectory of \method{} for a multi-hop preference inference task.
Given a cross-site query requesting an article on a regularly followed topic alongside a piece from a preferred category, \method{} retrieves two historical preferences: \texttt{Topic: Space} on the earth.com platform and \texttt{Category: Science} on the National Geographic website.
The agent initiates a two-hop trajectory by first opening a space-themed article (Steps 1--4).
Subsequently, the agent switches to the National Geographic domain and navigates through the Science section (Steps 5--7).
The task successfully concludes when the agent lands on a matching science article (Step 8).
}
\label{fig:qualititative_mh_pref}
\end{figure*}

\begin{figure*}[t]
    \centering
        \includegraphics[width=0.73\textwidth]{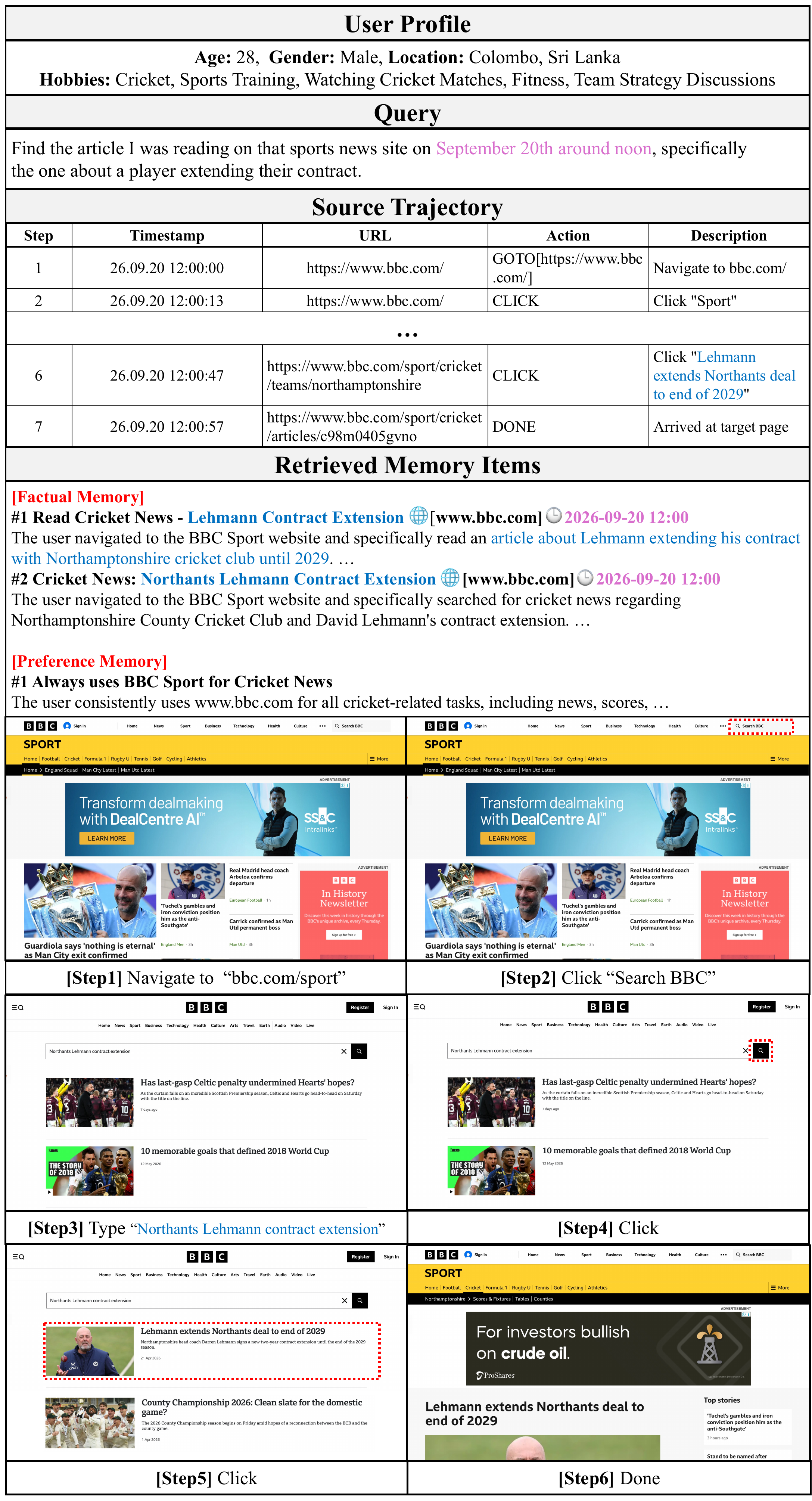}
        \caption{Execution trajectory of \method{} for a single-hop episodic grounding task.
Given a temporally-referenced query regarding a sports news article viewed on September 20th about a player contract extension, \method{} retrieves the corresponding factual memory (timestamped 2026.09.20 12:00).
The retrieved memory details a cricket article concerning the Northants Lehmann contract extension.
The agent navigates through the search interface of the BBC Sport website (Steps 2--4) to successfully land on the exact historical article (Step 5).
}
\label{fig:qualititative_sh_epi}
\end{figure*}

\begin{figure*}[t]
    \centering
        \includegraphics[width=0.71\textwidth]{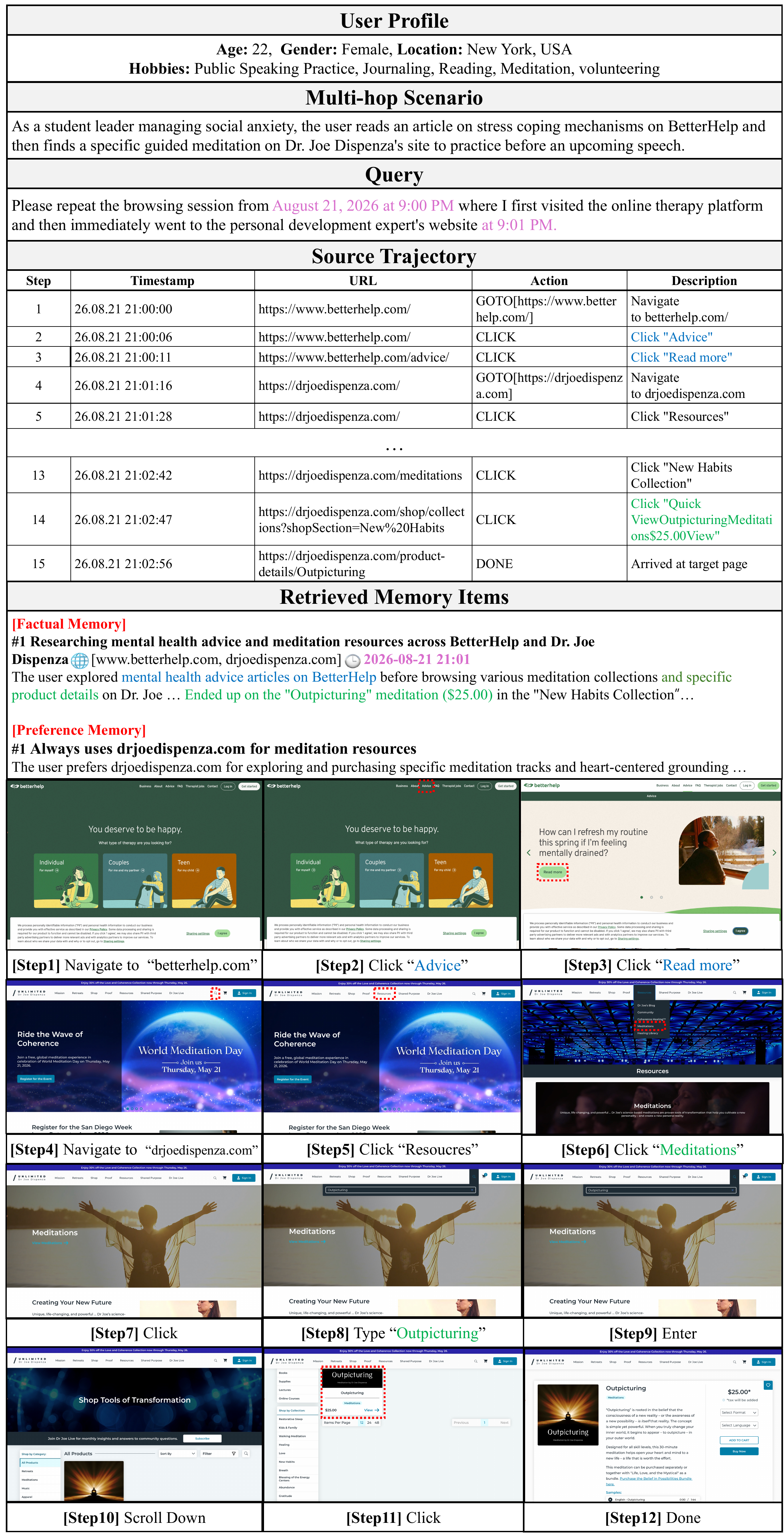}
        \caption{Execution trajectory of \method{} for a multi-hop episodic grounding task.
Given a temporally-referenced query to replay a past session across two distinct websites, \method{} accurately retrieves the factual memory timestamped for the exact specified date.
The agent navigates the BetterHelp domain to locate the historical stress-coping article and subsequently switches to the second platform to find the exact meditation product.
}
\label{fig:qualititative_mh_epi}
\end{figure*}

We present step-by-step execution trajectories across all four evaluation settings: single-hop and multi-hop preference inference, and single-hop and multi-hop episodic grounding.
Each trace illustrates the navigation actions of the agent and highlights the relevant memory utilized to fulfill the personalization task.
Specifically, Figure~\ref{fig:qualititative_sh_pref} and Figure~\ref{fig:qualititative_mh_pref} display the execution sequences for single-hop and multi-hop preference inference tasks, respectively.
Furthermore, Figure~\ref{fig:qualititative_sh_epi} and Figure~\ref{fig:qualititative_mh_epi} depict the corresponding trajectories for single-hop and multi-hop episodic grounding tasks.

\section{Prompt Templates}
\label{app:prompts}
This section outlines the prompt templates used to construct \benchmark and for the experimental evaluations.

 The prompts for \benchmark{} construction are organized by each stage of the construction pipeline: website curation (Figure~\ref{fig:domain_classification}), user profile generation and website assignment (Figures~\ref{fig:attribute_extraction}, ~\ref{fig:domain_assignment}, ~\ref{fig:website_selection}, ~\ref{fig:storyline_generation}, and ~\ref{fig:storyline_matching}), query template discovery (Figures~\ref{fig:template_discovery_1},
  ~\ref{fig:template_discovery_2}, and ~\ref{fig:template_discovery_3}), trajectory rollout (Figures~\ref{fig:trajectory_rollout_1} and ~\ref{fig:trajectory_rollout_2}), personalized query generation (Figures~\ref{fig:sh_preference_inference_query}, ~\ref{fig:mh_preference_inference_query}, ~\ref{fig:sh_episodic_grounding_query}, and ~\ref{fig:mh_episodic_grounding_query}), and verification (Figure~\ref{fig:trajectory_verification}).

\begin{figure*}[htbp]
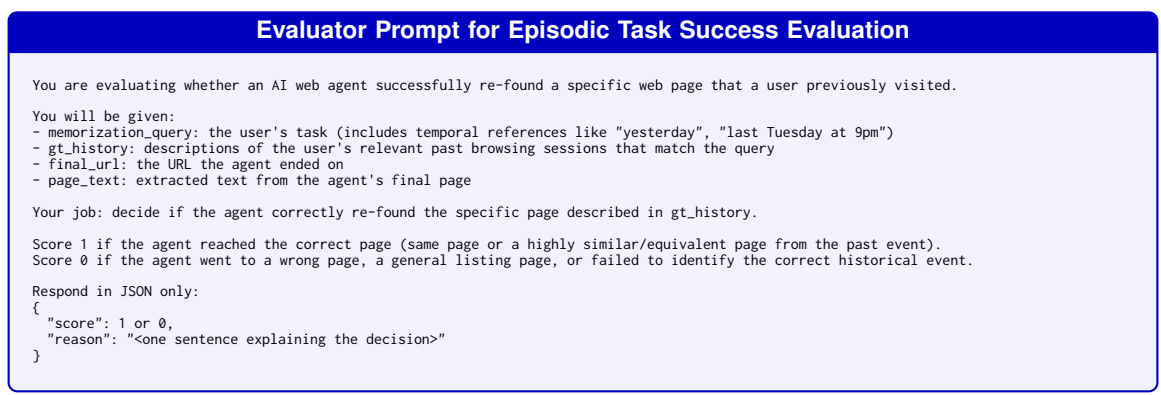

\centering
\begin{tcolorbox}[
  colback=blue!5!white,
  colframe=blue!75!black,
  width=0.95\textwidth,
  arc=1mm,
  boxrule=0.8pt,
  title=\textbf{Evaluator Prompt for Episodic Task Success Evaluation},
  colbacktitle=blue!75!black,
  coltitle=white,
  fonttitle=\bfseries\sffamily\small,
  halign title=center,
  top=2mm,
  bottom=2mm,
  left=2mm,
  right=2mm
]
\begin{lstlisting}[
  breaklines=true, 
  basicstyle=\ttfamily\tiny, 
  columns=fullflexible,
  xleftmargin=0pt,
  xrightmargin=0pt,
  aboveskip=1pt,
  belowskip=1pt,
  lineskip=-1pt,
  gobble=0,
  breakindent=0pt,
  breakautoindent=false
]
You are evaluating whether an AI web agent successfully re-found a specific web page that a user previously visited.

You will be given:
- memorization_query: the user's task (includes temporal references like "yesterday", "last Tuesday at 9pm")
- gt_history: descriptions of the user's relevant past browsing sessions that match the query
- final_url: the URL the agent ended on
- page_text: extracted text from the agent's final page

Your job: decide if the agent correctly re-found the specific page described in gt_history.

Score 1 if the agent reached the correct page (same page or a highly similar/equivalent page from the past event).
Score 0 if the agent went to a wrong page, a general listing page, or failed to identify the correct historical event.

Respond in JSON only:
{
  "score": 1 or 0,
  "reason": "<one sentence explaining the decision>"
}
\end{lstlisting}
\end{tcolorbox}
\caption{Evaluator prompt template for evaluating successful re-finding of a previously visited page.}
\label{fig:evaluator_task_success}
\end{figure*}

The prompts for \method{} are organized by each step: trajectory segmentation (Figure~\ref{fig:pacmem_trajectory_segmentation}), factual memory construction (Figures~\ref{fig:pacmem_episode_grouping} and ~\ref{fig:pacmem_factual_memory}), preference memory construction (Figure~\ref{fig:pacmem_preference_memory}), and memory retrieval (Figure~\ref{fig:pacmem_memory_retrieval}).
The system prompt for step-level browser action decision is provided in Figure~\ref{fig:system_prompt_base}.
For the baseline methods, we include the prompts for AWM (Figure~\ref{fig:awm_workflow_induction}) and ReasoningBank (Figures~\ref{fig:reasoningbank_success_prompt} and ~\ref{fig:reasoningbank_failed_prompt}).
Finally, we provide the LLM-as-judge prompts used in our evaluation pipeline (Figures~\ref{fig:evaluator_preference_score}, ~\ref{fig:evaluator_intent_score}, and ~\ref{fig:evaluator_task_success}).

\section{Use of Large Language Models}
\label{app:use_of_llms}
In accordance with the ACL Policy on Publication Ethics, we disclose our use of LLMs in writing this paper.
LLMs were used only as a writing assistant to improve grammatical correctness and readability.
They were not used to generate research ideas, paper structure, experimental results, or citations.
All model-generated suggestions were thoroughly reviewed, and the final text was written by the authors, who take full responsibility for the content.

\section{Artifact Licenses and Intended Use}
\label{app:licenses}
\paragraph{Used Artifacts.}
Tranco~\citep{Le_Pochat_2019} is used only for non-commercial academic research, consistent with its upstream sources' terms.
PersonaHub~\citep{ge2024scaling} is released under CC BY-NC-SA 4.0 and used only for non-commercial research.
Similarweb's publicly viewable category pages were consulted to identify per-category top websites and to adopt their two-level taxonomy.
We do not redistribute Similarweb's proprietary traffic data or ranking metrics.

\paragraph{Created Artifacts.}
\benchmark{} will be released under CC BY-NC-SA 4.0 to match PersonaHub's share-alike clause, and the \method{} code will be released under MIT. Both are intended for non-commercial research on personalized web agents, and the synthetic profiles inherit PersonaHub's research-use conditions.


\begin{figure*}[t]
    \centering
        \includegraphics[width=\textwidth]{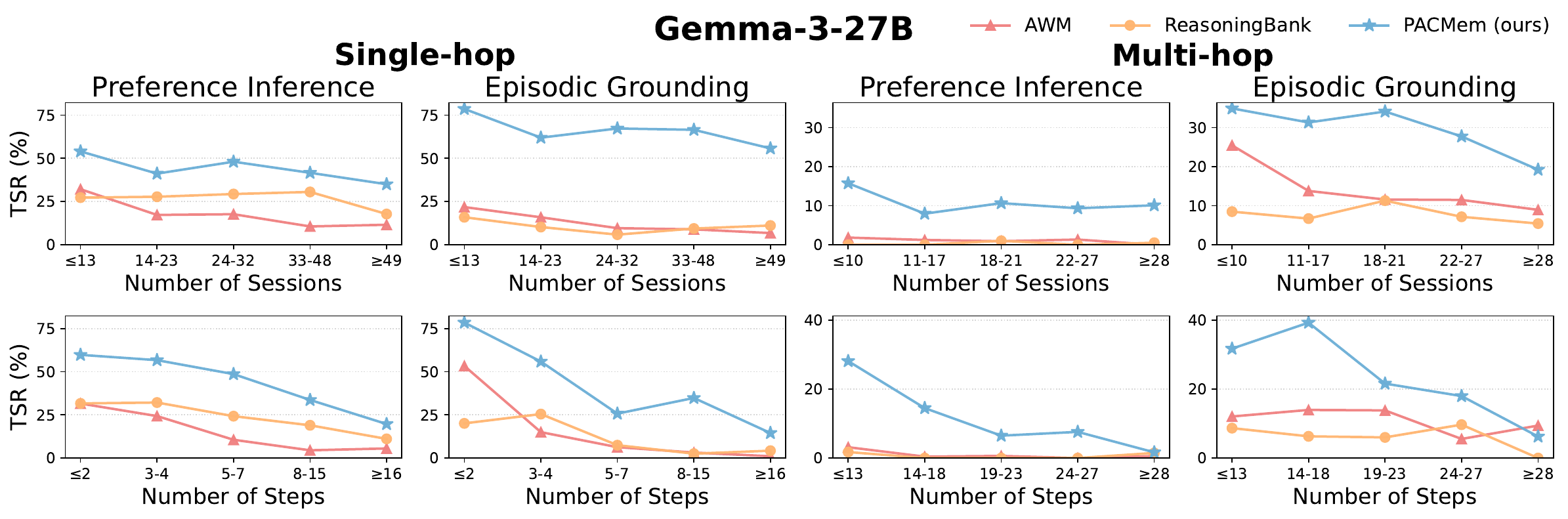}
        \caption{Analysis of TSR (\%) across varying numbers of sessions (top) and steps per task (bottom) on Gemma-3-27B.}
\label{fig:length_anslysis_gemma}
\end{figure*}

\begin{figure*}[t]
    \centering
        \includegraphics[width=\textwidth]{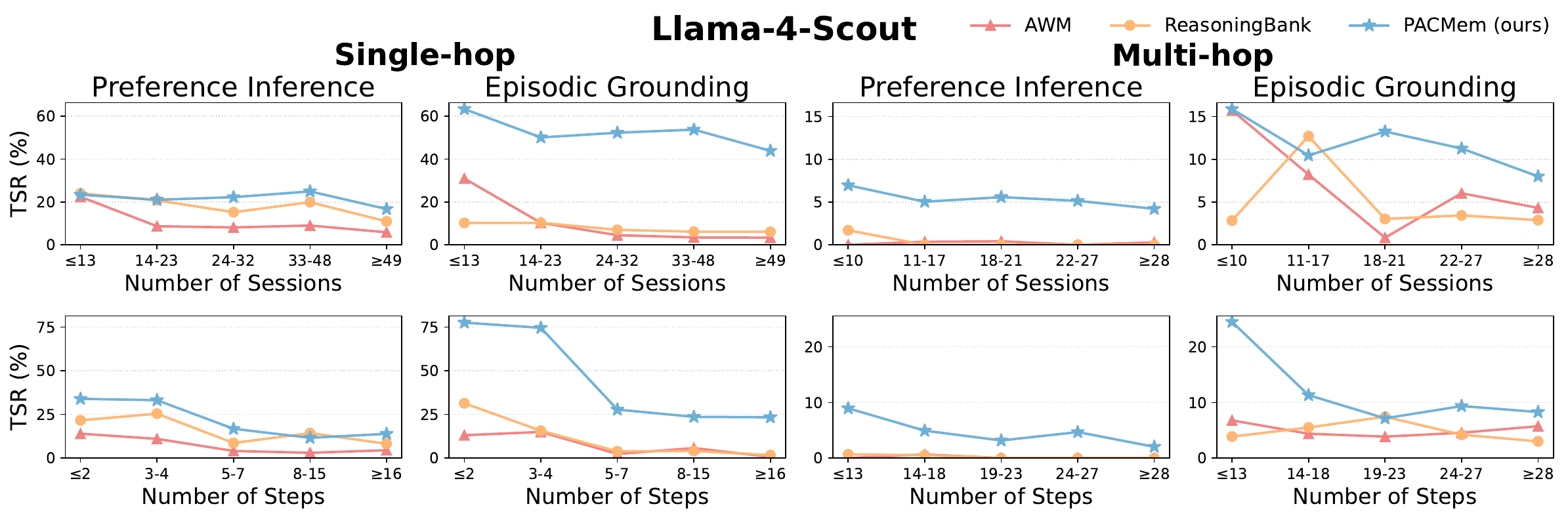}
        \caption{Analysis of TSR (\%) across varying numbers of sessions (top) and steps per task (bottom) on Llama-4-Scout.}
\label{fig:length_anslysis_llama}
\end{figure*}


\clearpage

\begin{table*}[t]
\centering
\small
\renewcommand{\arraystretch}{1.2}
\begin{tabular}{@{}p{0.16\textwidth}p{0.79\textwidth}@{}}
\toprule
\multicolumn{2}{c}{\textbf{Case 1: Preferred Artist Not Applied (Navigation Failure)}} \\
\midrule
\textbf{User query} & ``Find a concert by an artist I usually follow at a venue in my area and view its details.'' \\
\midrule
\textbf{Gold preference} & Artist\_Name = \textcolor{blue}{The Weeknd} \\
\midrule
\textbf{Retrieved memory} & \textbf{Item 1:} \textit{``Frequent interest in \textcolor{blue}{The Weeknd} concerts: The user repeatedly searches for concert information specifically related to the artist \textcolor{blue}{The Weeknd} across different locations.''}\newline\newline \textbf{Item 2:} \textit{``View \textcolor{blue}{The Weeknd} artist page: The user navigated to the artists section of Songkick, selected \textcolor{blue}{The Weeknd}, and viewed the artist's page. ...''}\newline\newline ... \\
\midrule
\textbf{Explored page} & songkick.com/festivals/3776908-\textcolor{red}{lee-seung-yoon-concert}-bagg/id/43040707-\textcolor{red}{lee-seung-yoon-concert}-{}-2026 \\
\midrule
\textbf{Gold page} & songkick.com/concerts/42774409-\textcolor{blue}{weeknd}-at-estadio-gnp-seguros \\
\bottomrule
\end{tabular}
\caption{A navigation failure case on preference inference, in which the correct preference is retrieved into the web agent's context but never applied during navigation. The retrieved memory explicitly states the target artist, and the web agent competently reaches a concert-details page (IS $=1$), but the page is a concert of a different artist, Lee Seung Yoon (PS $=0$).}
\label{tab:case_study_1}
\end{table*}

\begin{table*}[t]
\centering
\small
\renewcommand{\arraystretch}{1.2}
\begin{tabular}{@{}p{0.16\textwidth}p{0.79\textwidth}@{}}
\toprule
\multicolumn{2}{c}{\textbf{Case 2: Preferred Division Not Applied (Navigation Failure)}} \\
\midrule
\textbf{User query} & ``Find the tournament in the type I usually follow and view its details.'' \\
\midrule
\textbf{Gold preference} & Event\_Type = \textcolor{blue}{Policy Debate} \\
\midrule
\textbf{Retrieved memory} & \textbf{Item 1:} \textit{``Prefers viewing \textcolor{blue}{Policy Debate} division details: The user often drills down into specific tournaments to view the `\textcolor{blue}{Policy Debate}' or `\textcolor{blue}{Open Policy}' division pages.''}\newline\newline \textbf{Item 2:} \textit{``Always uses tabroom.com for debate tournament search: The user exclusively uses www.tabroom.com to find, browse, and view details for speech and debate tournaments.''}\newline\newline ... \\
\midrule
\textbf{Explored page} & www.tabroom.com/index/tourn/events.mhtml?tourn\_id=38772 \enspace (a \textcolor{red}{Congress debate} event page) \\
\midrule
\textbf{Gold page} & www.tabroom.com/index/tourn/events.mhtml?event\_id=341756\&tourn\_id=36419 \\
\bottomrule
\end{tabular}
\caption{A navigation failure case on preference inference, in which the correct preference is retrieved into the web agent's context but never applied during navigation. The retrieved memory explicitly names the preferred division, and the web agent reaches a tournament events page (IS $=1$), but the page shows an event of a different division, Congress debate (PS $=0$).}
\label{tab:case_study_2}
\end{table*}

\clearpage

\begin{table*}[t]
\centering
\small
\renewcommand{\arraystretch}{1.2}
\begin{tabular}{@{}p{0.16\textwidth}p{0.79\textwidth}@{}}
\toprule
\multicolumn{2}{c}{\textbf{Case 3: Retrieved but Not Re-reached (Navigation Failure)}} \\
\midrule
\textbf{User query} & ``Find the flight search I did on that airline booking site on January 26th at 11 AM, which was the third time I looked up an international destination that year.'' \\
\midrule
\textbf{Gold episode} & westjet.com/en-ca/destinations/discover/london-united-kingdom, visited \textcolor{blue}{2026-01-26 11:00} \\
\midrule
\textbf{Retrieved memory} & \textbf{Item 1:} \textit{``Westjet Flight Search: London, UK''} (\textcolor{blue}{2026-01-26 11:00}, final page: \textcolor{blue}{westjet.com/en-ca/destinations/discover/london-united-kingdom})\newline\newline ... \\
\midrule
\textbf{Explored page} & www.westjet.com/en-ca \enspace (\textcolor{red}{the site homepage}) \\
\midrule
\textbf{Gold page} & www.westjet.com/en-ca/destinations/discover/\textcolor{blue}{london-united-kingdom} \\
\bottomrule
\end{tabular}
\caption{A navigation failure case on episodic grounding. The correct entry, including the exact target URL and timestamp, is retrieved and injected into the web agent's context (ERR $=1$), yet the web agent wanders through search and menu pages and ends on the homepage after exhausting its step budget (TSR $=0$).}
\label{tab:case_study_4}
\end{table*}

\begin{table*}[t]
\centering
\small
\renewcommand{\arraystretch}{1.2}
\begin{tabular}{@{}p{0.16\textwidth}p{0.79\textwidth}@{}}
\toprule
\multicolumn{2}{c}{\textbf{Case 4: Relevant Entry Not Retrieved (Memory Failure)}} \\
\midrule
\textbf{User query} & ``Find the article I was reading on that how-to guide site on July 16th around 8 PM.'' \\
\midrule
\textbf{Gold episode} & wikihow.com/Learn-Faster, visited \textcolor{blue}{2026-07-16 20:00} \\
\midrule
\textbf{Retrieved memory} & \textbf{Item 1:} \textit{``Browsed wikiHow for Philosophy information''} (\textcolor{red}{2026-11-16}, final page: wikihow.com/Understand-Yourself)\newline\newline \textbf{Item 2:} \textit{``Read Neuroscience Article on Nature.com''} (\textcolor{red}{2026-01-19})\newline\newline \textbf{Item 3:} \textit{``Read Neuroscience Article on Nature.com''} (\textcolor{red}{2026-10-05}) \\
\midrule
\textbf{Unretrieved entry} & \textit{``How to Study - Wikihow Exploration''} (\textcolor{blue}{2026-07-16 20:00}, final page: \textcolor{blue}{wikihow.com/Learn-Faster}), which was present in the memory pool but not ranked into the retrieved set \\
\midrule
\textbf{Explored page} & www.wikihow.com/\textcolor{red}{Understand-Yourself} \\
\midrule
\textbf{Gold page} & www.wikihow.com/\textcolor{blue}{Learn-Faster} \\
\bottomrule
\end{tabular}
\caption{A memory failure case on episodic grounding. The memory pool contains the correct entry for the referenced session, but it is not retrieved (ERR $=0$). None of the retrieved entries matches the time in the query. The web agent therefore follows a wrong same-site entry and lands on a plausible but incorrect article (TSR $=0$).}
\label{tab:case_study_3}
\end{table*}

\clearpage
\onecolumn
\begin{longtable}{p{0.20\textwidth} p{0.24\textwidth} c p{0.20\textwidth}}
\toprule
\textbf{Domain} & \textbf{Subdomain} & \textbf{\# Websites} & \textbf{Websites} \\
\midrule
\endfirsthead
\multicolumn{4}{c}{\textit{(continued from previous page)}} \\
\toprule
\textbf{Domain} & \textbf{Subdomain} & \textbf{\# Websites} & \textbf{Websites} \\
\midrule
\endhead
\midrule
\multicolumn{4}{r}{\textit{(continued on next page)}} \\
\endfoot
\bottomrule
\endlastfoot
\multirow{17}{=}{\textbf{Science And Education}} & Education & 8 & \makecell[tl]{coursera \\ starfall \\ wordwall} \\
\cmidrule(lr){2-4}
 & History & 7 & \makecell[tl]{annefrank \\ westpoint \\ oerproject} \\
\cmidrule(lr){2-4}
 & Literature & 5 & \makecell[tl]{poetrysociety \\ bookishwayfarer \\ dunenovels} \\
\cmidrule(lr){2-4}
 & Philosophy & 4 & \makecell[tl]{safe \\ thepointmag \\ jim} \\
\cmidrule(lr){2-4}
 & Science And Education & 4 & \makecell[tl]{springer \\ nature \\ elsevier} \\
\cmidrule(lr){2-4}
 & Universities And Colleges & 4 & \makecell[tl]{upenn \\ yale \\ nyu} \\
\cmidrule(lr){2-4}
 & Weather & 4 & \makecell[tl]{wunderground \\ noaa \\ theweathernetwork} \\
\cmidrule(lr){2-4}
 & Biology & 3 & \makecell[tl]{hubermanlab \\ asm \\ physiology} \\
\cmidrule(lr){2-4}
 & Chemistry & 3 & \makecell[tl]{rsc \\ acs \\ cas} \\
\cmidrule(lr){2-4}
 & Environmental Science & 3 & \makecell[tl]{ucanr \\ noharm \\ earth} \\
\cmidrule(lr){2-4}
 & Public Records And Directories & 3 & \makecell[tl]{archivists \\ faqs \\ doxpop} \\
\cmidrule(lr){2-4}
 & Earth Sciences & 2 & \makecell[tl]{copernicus \\ gisgeography} \\
\cmidrule(lr){2-4}
 & Grants Scholarships And Financial Aid & 2 & \makecell[tl]{bold \\ scholarships} \\
\cmidrule(lr){2-4}
 & Libraries And Museums & 2 & \makecell[tl]{senecapolytechnic \\ bibliocommons} \\
\cmidrule(lr){2-4}
 & Math & 2 & \makecell[tl]{overleaf \\ wolfram} \\
\cmidrule(lr){2-4}
 & Physics & 2 & \makecell[tl]{alphaxiv \\ smodin} \\
\cmidrule(lr){2-4}
 & Social Sciences & 2 & \makecell[tl]{repec \\ tabroom} \\
\midrule
\multirow{10}{=}{\textbf{Health}} & Mental Health & 7 & \makecell[tl]{betterhelp \\ therapyappointment \\ psychologytoday} \\
\cmidrule(lr){2-4}
 & Alternative And Natural Medicine & 4 & \makecell[tl]{chirotouch \\ drjoedispenza \\ biogena} \\
\cmidrule(lr){2-4}
 & Geriatric And Aging Care & 4 & \makecell[tl]{agingcare \\ 10toptips \\ ncoa} \\
\cmidrule(lr){2-4}
 & Nutrition Diets And Fitness & 4 & \makecell[tl]{roguefitness \\ weightwatchers \\ thegymgroup} \\
\cmidrule(lr){2-4}
 & Public Health And Safety & 4 & \makecell[tl]{healthdata \\ nremt \\ nfpa} \\
\cmidrule(lr){2-4}
 & Addictions & 3 & \makecell[tl]{lightningstep \\ abby \\ blocksite} \\
\cmidrule(lr){2-4}
 & Mens Health & 2 & \makecell[tl]{movember \\ issm} \\
\cmidrule(lr){2-4}
 & Womens Health & 2 & \makecell[tl]{helloclue \\ volleyballworld} \\
\cmidrule(lr){2-4}
 & Health & 1 & \makecell[tl]{health} \\
\cmidrule(lr){2-4}
 & Medicine & 1 & \makecell[tl]{nyp} \\
\midrule
\multirow{13}{=}{\textbf{Sports}} & Fantasy Sports & 6 & \makecell[tl]{prizepicks \\ statmuse \\ basketballmonster} \\
\cmidrule(lr){2-4}
 & Running & 3 & \makecell[tl]{raceroster \\ theconqueror \\ nyrr} \\
\cmidrule(lr){2-4}
 & Soccer & 3 & \makecell[tl]{onefootball \\ flashscore \\ beinsports} \\
\cmidrule(lr){2-4}
 & American Football & 2 & \makecell[tl]{elevenwarriors \\ seahawks} \\
\cmidrule(lr){2-4}
 & Baseball & 2 & \makecell[tl]{mlbtraderumors \\ sny} \\
\cmidrule(lr){2-4}
 & Basketball & 2 & \makecell[tl]{nbcsportsbayarea \\ tankathon} \\
\cmidrule(lr){2-4}
 & Golf & 2 & \makecell[tl]{golfgenius \\ golf} \\
\cmidrule(lr){2-4}
 & Martial Arts & 2 & \makecell[tl]{tapology \\ bjpenn} \\
\cmidrule(lr){2-4}
 & Sports & 2 & \makecell[tl]{strava \\ yahoo} \\
\cmidrule(lr){2-4}
 & Climbing & 1 & \makecell[tl]{skimo} \\
\cmidrule(lr){2-4}
 & Fishing & 1 & \makecell[tl]{tides4fishing} \\
\cmidrule(lr){2-4}
 & Rugby & 1 & \makecell[tl]{ulster} \\
\cmidrule(lr){2-4}
 & Winter Sports & 1 & \makecell[tl]{onthesnow} \\
\midrule
\multirow{9}{=}{\textbf{Computers Electronics And Technology}} & Social Networks And Online Communities & 5 & \makecell[tl]{nextdoor \\ discord \\ slack} \\
\cmidrule(lr){2-4}
 & Computers Electronics And Technology & 3 & \makecell[tl]{moloco \\ tinyurl \\ okta} \\
\cmidrule(lr){2-4}
 & File Sharing And Hosting & 3 & \makecell[tl]{dropbox \\ sharefile \\ imgur} \\
\cmidrule(lr){2-4}
 & Graphics Multimedia And Web Design & 3 & \makecell[tl]{heygen \\ deevid \\ weebly} \\
\cmidrule(lr){2-4}
 & Computer Security & 2 & \makecell[tl]{tryhackme \\ bitdefender} \\
\cmidrule(lr){2-4}
 & Web Hosting And Domain Names & 2 & \makecell[tl]{amazon \\ miraheze} \\
\cmidrule(lr){2-4}
 & Advertising Networks & 1 & \makecell[tl]{revcontent} \\
\cmidrule(lr){2-4}
 & Computer Hardware & 1 & \makecell[tl]{raspberrypi} \\
\cmidrule(lr){2-4}
 & Programming And Developer Software & 1 & \makecell[tl]{office} \\
\midrule
\multirow{7}{=}{\textbf{Arts And Entertainment}} & Visual Arts And Design & 5 & \makecell[tl]{line-of-action \\ artsy \\ deviantart} \\
\cmidrule(lr){2-4}
 & Books And Literature & 4 & \makecell[tl]{goodreads \\ poetryfoundation \\ crazymaplestudios} \\
\cmidrule(lr){2-4}
 & Animation And Comics & 3 & \makecell[tl]{webtoons \\ comicbook \\ marvel} \\
\cmidrule(lr){2-4}
 & Music & 3 & \makecell[tl]{buzzsprout \\ spotify \\ spreaker} \\
\cmidrule(lr){2-4}
 & Arts And Entertainment & 2 & \makecell[tl]{district \\ pagesix} \\
\cmidrule(lr){2-4}
 & Performing Arts & 2 & \makecell[tl]{cirquedusoleil \\ frontgatetickets} \\
\cmidrule(lr){2-4}
 & Tv Movies And Streaming & 1 & \makecell[tl]{primevideo} \\
\midrule
\multirow{5}{=}{\textbf{Business And Consumer Services}} & Online Marketing & 6 & \makecell[tl]{flodesk \\ gemius \\ beehiiv} \\
\cmidrule(lr){2-4}
 & Publishing And Printing & 6 & \makecell[tl]{acx \\ condenast \\ blurb} \\
\cmidrule(lr){2-4}
 & Business Services & 3 & \makecell[tl]{acuityscheduling \\ statista \\ dataannotation} \\
\cmidrule(lr){2-4}
 & Business And Consumer Services & 2 & \makecell[tl]{flexmls \\ pirateship} \\
\cmidrule(lr){2-4}
 & Marketing And Advertising & 2 & \makecell[tl]{innovid \\ wistia} \\
\midrule
\multirow{5}{=}{\textbf{Hobbies And Leisure}} & Camping Scouting And Outdoors & 7 & \makecell[tl]{chicagoparkdistrict \\ reserveamerica \\ padi} \\
\cmidrule(lr){2-4}
 & Antiques And Collectibles & 3 & \makecell[tl]{providentmetals \\ jmbullion \\ radiomuseum} \\
\cmidrule(lr){2-4}
 & Photography & 3 & \makecell[tl]{pbase \\ 500px \\ befunky} \\
\cmidrule(lr){2-4}
 & Crafts & 2 & \makecell[tl]{thesprucecrafts \\ stitchfiddle} \\
\cmidrule(lr){2-4}
 & Ancestry And Genealogy & 1 & \makecell[tl]{wikitree} \\
\midrule
\multirow{5}{=}{\textbf{Law And Government}} & Government & 5 & \makecell[tl]{state \\ epa \\ ny} \\
\cmidrule(lr){2-4}
 & Legal & 3 & \makecell[tl]{aclu \\ justice \\ plusgrade} \\
\cmidrule(lr){2-4}
 & National Security & 3 & \makecell[tl]{airforce \\ cisa \\ fas} \\
\cmidrule(lr){2-4}
 & Immigration And Visas & 2 & \makecell[tl]{joinsherpa \\ visajourney} \\
\cmidrule(lr){2-4}
 & Law Enforcement And Protective Services & 2 & \makecell[tl]{themarshallproject \\ crimestoppers-uk} \\
\midrule
\multirow{5}{=}{\textbf{Community And Society}} & Philanthropy & 7 & \makecell[tl]{compassion \\ torontomu \\ dewatermark} \\
\cmidrule(lr){2-4}
 & Faith And Beliefs & 4 & \makecell[tl]{bible \\ biblehub \\ horoscope} \\
\cmidrule(lr){2-4}
 & Lgbtq & 2 & \makecell[tl]{hrc \\ xtramagazine} \\
\cmidrule(lr){2-4}
 & Decease & 1 & \makecell[tl]{arbormemorial} \\
\cmidrule(lr){2-4}
 & Holidays And Seasonal Events & 1 & \makecell[tl]{print-a-calendar} \\
\midrule
\multirow{3}{=}{\textbf{Reference Materials}} & Reference Materials & 7 & \makecell[tl]{wikihow \\ manualslib \\ worldpopulationreview} \\
\cmidrule(lr){2-4}
 & Dictionaries And Encyclopedias & 3 & \makecell[tl]{yourdictionary \\ oxfordlearnersdictionaries \\ mapy} \\
\cmidrule(lr){2-4}
 & Maps & 2 & \makecell[tl]{onefivenine \\ worldatlas} \\
\midrule
\multirow{4}{=}{\textbf{Games}} & Board And Card Games & 4 & \makecell[tl]{scryfall \\ solitaired \\ boardgamearena} \\
\cmidrule(lr){2-4}
 & Video Games Consoles And Accessories & 3 & \makecell[tl]{op \\ twitch \\ hytale} \\
\cmidrule(lr){2-4}
 & Puzzles And Brainteasers & 2 & \makecell[tl]{lumosity \\ crosswordsolver} \\
\cmidrule(lr){2-4}
 & Roleplaying Games & 1 & \makecell[tl]{gmbinder} \\
\midrule
\multirow{1}{=}{\textbf{News And Media}} & News And Media & 10 & \makecell[tl]{bbc \\ time \\ techcrunch} \\
\midrule
\multirow{4}{=}{\textbf{Travel And Tourism}} & Travel And Tourism & 4 & \makecell[tl]{lonelyplanet \\ travelandleisure \\ carnival} \\
\cmidrule(lr){2-4}
 & Air Travel & 3 & \makecell[tl]{westjet \\ flightaware \\ qantas} \\
\cmidrule(lr){2-4}
 & Accommodation And Hotels & 1 & \makecell[tl]{accor} \\
\cmidrule(lr){2-4}
 & Tourist Attractions & 1 & \makecell[tl]{sixflags} \\
\midrule
\multirow{3}{=}{\textbf{Jobs And Career}} & Human Resources & 4 & \makecell[tl]{16personalities \\ adobe \\ ashbyhq} \\
\cmidrule(lr){2-4}
 & Jobs And Career & 2 & \makecell[tl]{builtin \\ charlotte} \\
\cmidrule(lr){2-4}
 & Jobs And Employment & 2 & \makecell[tl]{appcast \\ ibps} \\
\midrule
\multirow{6}{=}{\textbf{Lifestyle}} & Fashion And Apparel & 2 & \makecell[tl]{harpersbazaar \\ vogue} \\
\cmidrule(lr){2-4}
 & Weddings & 2 & \makecell[tl]{pic-time \\ passgallery} \\
\cmidrule(lr){2-4}
 & Beauty And Cosmetics & 1 & \makecell[tl]{byrdie} \\
\cmidrule(lr){2-4}
 & Childcare & 1 & \makecell[tl]{parents} \\
\cmidrule(lr){2-4}
 & Gifts And Flowers & 1 & \makecell[tl]{fromyouflowers} \\
\cmidrule(lr){2-4}
 & Lifestyle & 1 & \makecell[tl]{thecut} \\
\midrule
\multirow{3}{=}{\textbf{Finance}} & Investing & 3 & \makecell[tl]{investopedia \\ etrade \\ ml} \\
\cmidrule(lr){2-4}
 & Accounting And Auditing & 2 & \makecell[tl]{xero \\ jacksonhewitt} \\
\cmidrule(lr){2-4}
 & Banking Credit And Lending & 2 & \makecell[tl]{squareup \\ chase} \\
\midrule
\multirow{2}{=}{\textbf{Food And Drink}} & Cooking And Recipes & 4 & \makecell[tl]{cookpad \\ recipetineats \\ simplyrecipes} \\
\cmidrule(lr){2-4}
 & Beverages & 2 & \makecell[tl]{vivino \\ livcheers} \\
\midrule
\multirow{4}{=}{\textbf{Heavy Industry And Engineering}} & Heavy Industry And Engineering & 2 & \makecell[tl]{altec \\ frazier} \\
\cmidrule(lr){2-4}
 & Architecture & 1 & \makecell[tl]{ncarb} \\
\cmidrule(lr){2-4}
 & Chemical Industry & 1 & \makecell[tl]{fmc} \\
\cmidrule(lr){2-4}
 & Construction And Maintenance & 1 & \makecell[tl]{procore} \\
\midrule
\multirow{2}{=}{\textbf{E Commerce And Shopping}} & Tickets & 3 & \makecell[tl]{ticketnew \\ songkick \\ luma} \\
\cmidrule(lr){2-4}
 & E Commerce And Shopping & 1 & \makecell[tl]{bigcartel} \\
\midrule
\multirow{2}{=}{\textbf{Home And Garden}} & Gardening & 2 & \makecell[tl]{diyeverywhere \\ epicgardening} \\
\cmidrule(lr){2-4}
 & Interior Design & 1 & \makecell[tl]{homedit} \\
\midrule
\multirow{1}{=}{\textbf{Pets And Animals}} & Animals & 3 & \makecell[tl]{exo-terra \\ campbowwow \\ petsnowy} \\
\midrule
\multirow{1}{=}{\textbf{Ai Chatbots And Tools}} & Ai Chatbots And Tools & 3 & \makecell[tl]{polybuzz \\ character \\ talkie-ai} \\
\midrule
\multirow{2}{=}{\textbf{Vehicles}} & Boats & 2 & \makecell[tl]{rolexsydneyhobart \\ boatinternational} \\
\cmidrule(lr){2-4}
 & Aviation & 1 & \makecell[tl]{avherald} \\
\end{longtable}
\addtocounter{table}{-1}
\captionof{table}{Domains, subdomains, and websites comprising the proposed benchmark. Websites within each sub-domain are ranked by query-generation frequency, capped at three.}
\label{tab:dataset_detail_1}

\clearpage

\clearpage
\prompttbox{Classifier Prompt for Website Domain Classification}
           {prompts/benchmark/domain_classification.txt}
           {Prompt for the LLM that classifies each curated website into the (domain, subdomain) structure used during Website Curation.}
           {domain_classification}

\prompttbox{Attribute Extraction Prompt for User Profile Generation}
           {prompts/benchmark/attribute_extraction.txt}
           {Prompt for the LLM that extracts structured attributes from a PersonaHub persona during User Profile Generation.}
           {attribute_extraction}

\prompttbox{Domain Assignment Prompt for Profile-Website Assignment}
           {prompts/benchmark/domain_assignment.txt}
           {Prompt for the LLM that assigns relevant domains and subdomains to a user profile based on the persona.}
           {domain_assignment}

\prompttbox{Website Selection Prompt for Profile-Website Assignment}
           {prompts/benchmark/website_selection.txt}
           {Prompt for selecting a website within each assigned subdomain that best fits the user profile.}
           {website_selection}
\prompttbox{Cross-Site Storyline Generation Prompt for Multi-Hop Assignment}
           {prompts/benchmark/storyline_generation.txt}
           {Prompt for the LLM that generates a pool of cross-site storylines, each spanning two or three subdomains under a unified goal.}
           {storyline_generation}
           
\prompttbox{Cross-Site Storyline Matching Prompt for Multi-Hop Assignment}
         {prompts/benchmark/storyline_matching.txt}
         {Prompt for matching cross-site storylines to each user profile to form multi-hop assignments.}
         {storyline_matching}
        
\prompttbox{Web Agent Prompt for Query Template Discovery (Part 1/3): Agent Role and Constraints 1--4}
         {prompts/benchmark/template_discovery_part1.txt}
         {Web agent prompt for query template discovery: agent role and the first four template constraints.}
         {template_discovery_1}

\prompttbox{Web Agent Prompt for Query Template Discovery (Part 2/3): Preference Observability Constraint}
         {prompts/benchmark/template_discovery_part2.txt}
         {Web agent prompt for query template discovery: the preference observability constraint with its three qualifying interaction channels.}
         {template_discovery_2}

\prompttbox{Web Agent Prompt for Query Template Discovery (Part 3/3): Output Specification}
         {prompts/benchmark/template_discovery_part3.txt}
         {Web agent prompt for query template discovery: template formatting rules, navigation actions, and output format.}
         {template_discovery_3}
         
\prompttbox{Web Agent Prompt for Trajectory Rollout (Part 1/2): Initial Trajectory}
             {prompts/benchmark/trajectory_rollout_part1.txt}
             {Prompt for executing an instantiated query template with the preference expressed through site-native channels.}
             {trajectory_rollout_1}

  \prompttbox{Web Agent Prompt for Trajectory Rollout (Part 2/2): Additional Trajectories under Fixed Preference}
             {prompts/benchmark/trajectory_rollout_part2.txt}
             {Additions for subsequent rollouts that require each new target to differ from prior ones.}
             {trajectory_rollout_2}

\prompttbox{Query Generation Prompt for Single-Hop Preference Inference}
         {prompts/benchmark/sh_preference_inference_query.txt}
         {Prompt for rewriting a single-hop query into an underspecified request that hides preference and target values.}
         {sh_preference_inference_query}
\prompttbox{Query Generation Prompt for Multi-Hop Preference Inference}
         {prompts/benchmark/mh_preference_inference_query.txt}
         {Prompt for rewriting a multi-hop query into a single underspecified sentence covering the entire cross-site task.}
         {mh_preference_inference_query}
\prompttbox{Query Generation Prompt for Single-Hop Episodic Grounding}
         {prompts/benchmark/sh_episodic_grounding_query.txt}
         {Prompt for rewriting a single-hop query with temporal or event-based cues referencing a specific past trajectory.}
         {sh_episodic_grounding_query}

\prompttbox{Query Generation Prompt for Multi-Hop Episodic Grounding}
         {prompts/benchmark/mh_episodic_grounding_query.txt}
         {Prompt for rewriting a multi-hop query as a single sentence with temporal or event-based cues referencing the cross-site session.}
         {mh_episodic_grounding_query}

\prompttbox{Verifier Prompt for Trajectory Screening}
         {prompts/benchmark/trajectory_verification.txt}
         {Prompt for screening each trajectory on preference consistency, task completion, and preference observability.}
         {trajectory_verification}

\prompttbox{Prompt for Trajectory Segmentation in \method{}}
         {prompts/method/trajectory_segmentation.txt}
         {Prompt for segmenting a continuous browsing stream into atomic, goal-oriented episodes by detecting semantic shifts.}
         {pacmem_trajectory_segmentation}
\prompttbox{Prompt for Episode Grouping in \method{}}
         {prompts/method/episode_grouping.txt}
         {Prompt for deciding whether two adjacent browsing sessions belong to the same high-level goal and should be grouped into one factual memory entry.}
         {pacmem_episode_grouping}
\prompttbox{Prompt for Factual Memory Distillation in \method{}}
         {prompts/method/factual_memory_distillation.txt}
         {Prompt for distilling grouped episodes into a structured factual memory entry with a descriptive title and natural language summary.}
         {pacmem_factual_memory}
\prompttbox{Prompt for Preference Memory Construction in \method{}}
         {prompts/method/preference_memory.txt}
         {Prompt for extracting recurring behavioral patterns from factual memories.}
         {pacmem_preference_memory}
\prompttbox{Prompt for Memory Retrieval in \method{}}
         {prompts/method/memory_retrieval.txt}
         {Prompt for scoring candidate memory entries by semantic relevance, temporal alignment, and practical utility for the current task.}
         {pacmem_memory_retrieval}
\prompttbox{Web Agent System Prompt for Browser Action Decision}
         {prompts/experiment/system_prompt_base.txt}
         {Base system prompt for the LLM agent that produces the next browser action over the Browser-Use action space.}
         {system_prompt_base}
\prompttbox{Prompt for Workflow Induction in AWM}
         {prompts/experiment/awm_workflow_induction.txt}
         {Prompt used in AWM to extract reusable workflows from past trajectories.}
         {awm_workflow_induction}
\prompttbox{Prompt for Reasoning Pattern Distillation in ReasoningBank (Successful Trajectories)}
         {prompts/experiment/reasoningbank_success.txt}
         {Prompt used in ReasoningBank to distill generalizable reasoning patterns from successful execution traces.}
         {reasoningbank_success_prompt}
\prompttbox{Prompt for Reasoning Pattern Distillation in ReasoningBank (Failed Trajectories)}
         {prompts/experiment/reasoningbank_failed.txt}
         {Prompt used in ReasoningBank to distill generalizable reasoning patterns from failed execution traces.}  
         {reasoningbank_failed_prompt}
\prompttbox{Evaluator Prompt for
   Preference Score}
         {prompts/experiment/preference_score.txt}
         {Evaluator prompt for the Preference Score metric, judging whether the agent successfully applied the user's preferences.}
        {evaluator_preference_score}

\prompttbox{Evaluator Prompt for
Intent Score}
     {prompts/experiment/intent_score.txt}
     {Evaluator prompt for the Intent Score metric, judging task completion from the final landing page independent of personalization.}
  {evaluator_intent_score}

\end{document}